\documentclass[a4paper]{article}
\newif\ifblind
\blindfalse
\usepackage{geometry}

\usepackage{setspace}
\usepackage{float}
\usepackage{graphicx}
\usepackage{epstopdf}
\newif\ifarxiv
\arxivtrue
\ifarxiv\else
  \usepackage{xr-hyper} % cross-document refs to Supplementary.tex; must load before hyperref
\fi
\usepackage{hyperref}
\hypersetup{colorlinks=true, allcolors=black, urlcolor=blue}
\usepackage{standalone}
\usepackage{amsmath,amsfonts,amsthm,bm} % Math packages
\usepackage{color}

\usepackage{pdflscape}
\usepackage{authblk}
\usepackage{subcaption}
\usepackage[flushleft]{threeparttable}
\usepackage[comma,authoryear]{natbib}
\usepackage{enumitem}
\usepackage{xcolor}
\usepackage{soul}
\usepackage{makecell,booktabs}
\usepackage{caption,dcolumn}
\usepackage{verbatim} 
\newcolumntype{d}[1]{D{.}{.}{4}}

\usepackage{dsfont}
\usepackage{mathrsfs}

\usepackage{multirow}
\usepackage{array}
\newcolumntype{P}[1]{>{\centering\arraybackslash}p{#1}}

\usepackage [english]{babel}
\usepackage [autostyle, english = american]{csquotes}
\MakeOuterQuote{"}

\usepackage{mathtools}

\newcommand{\interior}[1]{%
 {\kern0pt#1}^{\mathrm{o}}%
}

\usepackage{algorithm,algpseudocode}
\usepackage[makeroom]{cancel}

\makeatletter
\newcommand*\bigcdot{\mathpalette\bigcdot@{.5}}
\newcommand*\bigcdot@[2]{\mathbin{\vcenter{\hbox{\scalebox{#2}{$\m@th#1\bullet$}}}}}
\makeatother

\makeatletter
\newcommand*{\addFileDependency}[1]{%
  \typeout{(#1)}%
  \@addtofilelist{#1}%
  \IfFileExists{#1}{}{\typeout{No file #1.}}%
}
\makeatother
\newcommand*{\myexternaldocument}[1]{%
  \externaldocument{#1}%
  \addFileDependency{#1.tex}%
  \addFileDependency{#1.aux}%
}
\ifarxiv\else\myexternaldocument{Supplementary}\fi
\usepackage{tabularx}

\title{\textbf{Prediction Uncertainty Controlled Computer Model Calibration}}

\ifblind
\author{}
\date{}
\else
\author[1,2]{Ryan Heejoon Byun}
\author[1,2]{Dayeon Yoon}
\author[1,2]{Won Chang\thanks{Corresponding Author, wonchang@snu.ac.kr}}
\date{}
\affil[1]{Department of Statistics, Seoul National University}
\affil[2]{Institute for Data Innovation in Science, Seoul National University}
\fi

\begin{document}

\maketitle

\begin{abstract}
Computer-model calibration is usually undertaken in service of prediction, yet neither Bayesian Kennedy–O'Hagan (KOH) calibration nor amortized simulation-based inference is built to control the predictive error of the calibrated simulator. When parameters are weakly identified or the data–model discrepancy is misspecified, the simulator-output bands induced by a nominal posterior level can be far from nominal. We propose a two-layer framework that targets predictive miscoverage directly. First, a conditional flow-matching model learns an amortized posterior for the calibration parameters, trained on simulator outputs perturbed with Gaussian-process discrepancies of varying structure. Second, conformal risk control arranges posterior draws for the target observation into nested candidate sets and selects the smallest one whose fixed-rank pointwise bands attain a prescribed time-averaged miscoverage on a synthetic calibration sample. Under exchangeability of calibration and target observations, the bands control expected miscoverage in finite samples. In a synthetic benchmark, the proposed bands are roughly 20\% the width of KOH–MCMC bands with miscoverage near the 10\% target. In calibrating an age-structured SEIR model to the 2010 Malawi measles outbreak, they are again sharper and near-nominal, and the weeks they fail to cover coincide with the asymmetric epidemic decline, exposing structural inadequacy of the mechanistic model.

\end{abstract}

\noindent%


\section{Introduction}
\label{sec:introduction}

Computer models---deterministic simulators of complex physical, biological, and engineering systems---are indispensable scientific instruments in fields ranging from epidemiology and climate science to hydrology and astrophysics. Their practical utility depends on \emph{calibration}: learning unknown simulator inputs from observations so that the calibrated simulator can support prediction, intervention design, and scientific inference \citep{kennedy2001bayesian,bayarri2007computer,Higdon2008,brynjarsdottir2014learning,plumlee2017bayesian}. These downstream tasks require more than accurate parameter point estimates; they require predictive uncertainty that is neither misleadingly narrow nor unnecessarily diffuse. For example, \citet{lim2025history} developed a more accurate method for estimating epidemic initial conditions but identified wide credible intervals and below-nominal empirical coverage as limitations requiring further work. Motivated by this concern, we focus on predictive uncertainty for simulator outputs: intervals should accurately identify scientifically relevant trajectories.

An increasingly attractive approach to calibration is to learn the inverse map from simulator outputs to input parameters. This inverse-model approach has been developed for high-dimensional dependent outputs using deep neural networks \citep{bhatnagar2022computer} and can be viewed within the broader framework of \emph{simulation-based inference} (SBI), in which simulations rather than a tractable likelihood provide the information needed for inference \citep{cranmer2020frontier}. Modern neural SBI methods train a conditional density estimator on simulated parameter--output pairs and thereby amortize the cost of inference: after a single training stage, the fitted model can rapidly return an approximate posterior for each new observation \citep{papamakarios2019sequential,papamakarios2016epsilonfree,greenberg2019automatic}, which makes SBI compelling for repeated, sequential, or time-sensitive calibration problems.

Amortization, however, does not by itself make posterior predictive uncertainty reliable. Under weak parameter identifiability, data--model discrepancy, or approximation error in the learned inverse map, some calibration parameters can be recovered imprecisely, and propagating the resulting parameter distribution through the simulator can produce predictive intervals that are excessively wide, too narrow, or otherwise poorly aligned with their nominal level \citep{hermans2022crisis}. Standard SBI procedures approximate parameter distributions; they do not control the predictive error of their simulator pushforwards. The conventional Kennedy--O'Hagan (KOH) Bayesian formulation \citep{kennedy2001bayesian} shares this predictive limitation while introducing a computational one: its posterior is commonly explored by Markov chain Monte Carlo (MCMC), which can be expensive and must generally be rerun for every new observation \citep{bhatnagar2022computer,cranmer2020frontier}. Moreover, nominal posterior probability levels and the repeated-sample coverage of simulator-output bands are different quantities, and parameter--discrepancy nonidentifiability or misspecification can further compromise inference \citep{brynjarsdottir2014learning,syring2019calibrating}. Figure~\ref{fig:uncalibrated-motivation} previews how KOH--MCMC and a na\"ive SBI construction can both yield conservative simulator-output bands for the same measles outbreak. We consequently take amortized SBI as our methodological starting point and use MCMC-based KOH calibration as a benchmark.

Our central contribution is to introduce \emph{prediction-accuracy-based conformal risk control} (CRC) into neural SBI. We combine CRC \citep{angelopoulos2024conformal} with flow-matching posterior estimation (FMPE) \citep{dax2023flowmatching}---a scalable neural posterior estimator based on continuous normalizing flows \citep{lipman2023flowmatching} with competitive performance on established SBI benchmarks \citep{dax2023flowmatching,lueckmann2021benchmark}. Rather than accepting a fixed posterior probability region, CRC selects the calibrated parameter set by directly evaluating a predictive miscoverage loss after parameter draws are propagated through the simulator, identifying the region of the parameter space most relevant for accurate prediction. Under exchangeability between the CRC calibration examples and a future target, the resulting simulator-output intervals control expected miscoverage in finite samples. This prediction-driven correction is related in spirit to flow-based conformal forecasting \citep{lee2026flowconformal}, but our target is the simulator pushforward of an amortized parameter posterior in a computer-model calibration problem. This prediction-driven calibration is practical only because the posterior is amortized: CRC requires an approximate posterior for each of hundreds of synthetic calibration observations conditioned on the target, which amounts to a few hundred ODE integrations with a trained flow but would demand as many separate MCMC runs under the standard KOH workflow, repeated for every new observation. Amortization is therefore not merely a computational convenience but the condition that makes prediction-accuracy-based risk control feasible.

We also address the discrepancy sensitivity of the inverse map. Following the learning-with-noise strategy of \citet{bhatnagar2022computer}, we train FMPE on simulator outputs contaminated by draws from a flexible heterogeneous Gaussian-process discrepancy family and generate the separate CRC calibration sample under the corresponding discrepancy mechanism. The framework therefore combines rapid amortized parameter inference with a second, prediction-focused layer that corrects the size of the simulator-output set, and it can be applied to SBI methods other than FMPE.

The remainder of the paper is organized as follows. Section~\ref{sec:background} reviews KOH and simulation-based calibration, conformal risk control, and flow matching. Section~\ref{sec:methods} develops the proposed framework. Section~\ref{sec:simulations} reports a simulation study using the synthetic model of \citet{bhatnagar2022computer}, Section~\ref{sec:applications} applies the framework to the Malawi measles outbreak, and Section~\ref{sec:discussion} concludes.

\section{Background}
\label{sec:background}

\subsection{Kennedy–O’Hagan (KOH) Calibration and Simulation-Based Inference}
\label{subsec:KOH-SBI}

The standard KOH calibration framework, proposed by \citet{kennedy2001bayesian} and developed in many subsequent works \citep[see, e.g.,][]{bayarri2007computer,Higdon2008,tuo2015efficient,plumlee2017bayesian,plumlee2019computer}, is based on the observation model
\begin{equation} \label{eq:KOH}
Z = Y(\theta) + \delta,
\end{equation}
where $\theta \in \Theta \subset \mathbb{R}^d$ denotes the calibration parameter vector; $Y(\theta) \in \mathbb{R}^T$ is the deterministic simulator output at $T$ discrete time points or spatial locations; $Z \in \mathbb{R}^T$ denotes the corresponding observational data; and $\delta \in \mathbb{R}^T$ is a \emph{data--model discrepancy} term capturing systematic mismatch between the simulator and reality, including observational errors. Given a prior $p(\theta)$ and the likelihood $p(Z \mid \theta)$ induced by \eqref{eq:KOH}, KOH calibration targets the posterior $p(\theta \mid Z) \propto p(Z \mid \theta)p(\theta)$, which quantifies \emph{parametric uncertainty}; propagating posterior draws through $Y(\theta)$ yields a posterior distribution for the simulator output. This forward formulation is statistically delicate: $\theta$ and the discrepancy can be nonidentifiable without careful prior specification or experimental design, and ignoring or misspecifying the discrepancy can bias inference on the physical parameters \citep{brynjarsdottir2014learning,tuo2015efficient,plumlee2017bayesian}. It is also computationally demanding, since a sufficiently long MCMC chain must be run for the observation at hand and repeated when a new observation arrives.

\begin{figure}[!tbp]
\centering
\begin{subfigure}{0.8\textwidth}
\centering
\includegraphics[width=\linewidth]{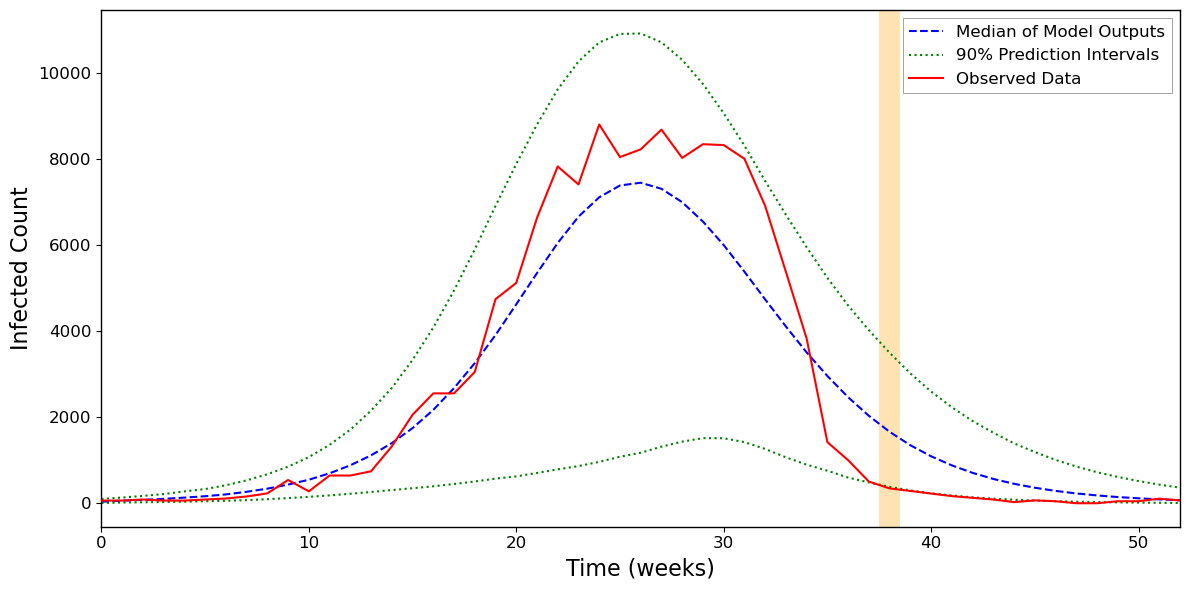}
\caption{KOH--MCMC posterior simulator-output band.}
\label{fig:seir-mcmc}
\end{subfigure}

\smallskip
\begin{subfigure}{0.8\textwidth}
\centering
\includegraphics[width=\linewidth]{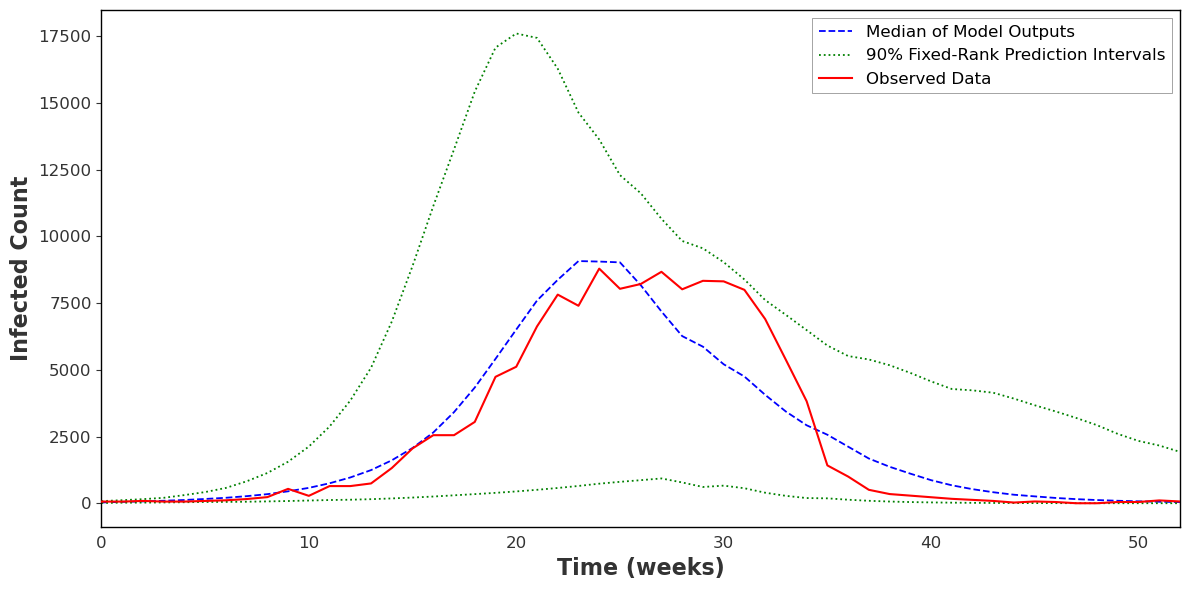}
\caption{Flow-matching SBI with fixed posterior-density retention and no CRC.}
\label{fig:naive-sbi}
\end{subfigure}
\caption{Predictive bands for the 2010 Malawi measles outbreak. Panel~(a) shows the nominal $90\%$ pointwise KOH--MCMC band: empirical miscoverage is $3.45\%$ on the evaluated weeks and average pointwise width is $3{,}526.89$ cases. Panel~(b) shows a flow-matching SBI band obtained by retaining the highest-density $90\%$ of posterior parameter draws: empirical miscoverage is $0\%$ and average width is $6{,}078.54$ cases. Blue dashed curves are medians of simulator outputs, green dotted curves are band limits, and red solid curves are observed weekly cases; orange shading marks missed observations when present. Miscoverage is evaluated only at weeks with at least $300$ observed cases.}
\label{fig:uncalibrated-motivation}
\end{figure}

A separate issue concerns the predictive meaning of the resulting bands. A nominal posterior probability level describes uncertainty under the assumed model; it does not, by itself, guarantee that a band covers observed outcomes at the same rate under repeated sampling \citep{syring2019calibrating}, and conventional KOH calibration contains no step that tunes simulator-output bands to a prespecified predictive-miscoverage loss. Figure~\ref{fig:seir-mcmc} illustrates this with the KOH--MCMC baseline for the Malawi measles outbreak analyzed in Section~\ref{sec:applications}. Its nominal $90\%$ pointwise band has empirical miscoverage $0.0345$ among weeks with at least $300$ reported cases (coverage of approximately $96.55\%$). The broad band, particularly around the epidemic peak, illustrates the cost in sharpness when posterior uncertainty is propagated without an explicit predictive-risk calibration step.

The inverse-model approach reverses the computational direction and learns calibration parameters directly from simulator outputs; \citet{bhatnagar2022computer}, for example, train a deep neural network on simulated parameter--output pairs for time-series calibration. More broadly, inference from an implicit simulator is described as \emph{simulation-based inference} \citep{cranmer2020frontier}. In neural posterior estimation, one simulates pairs $(\theta,x)$ from the joint model and fits a conditional density $q_\phi(\theta\mid x)$ to approximate $p(\theta\mid x)$ \citep{papamakarios2016epsilonfree,greenberg2019automatic}; the training cost is paid once, after which $q_\phi(\theta\mid Z)$ can be sampled for any new observation at negligible additional cost. SBI nevertheless changes the computational mechanism, not the predictive objective. If a component of $\theta$ is weakly identified, the learned conditional distribution is inaccurate, or the observed discrepancy differs from those represented during training, the approximate posterior may be displaced or overly dispersed; passing all of its draws through $Y(\theta)$ transfers that error to the output scale. A nominal credible region from KOH--MCMC and a nominal probability region from SBI therefore share a crucial limitation: neither is constructed to guarantee that the corresponding simulator-output interval attains its nominal predictive coverage \citep[see, e.g.,][]{hermans2022crisis,syring2019calibrating}.

Figure~\ref{fig:naive-sbi} makes the SBI limitation concrete using approximate posterior samples from a trained flow-matching model for the same Malawi outbreak. We retain the $90\%$ of generated parameter draws with the highest estimated conditional posterior density, pass them through the SEIR simulator, and form pointwise bands from the forward trajectories. The band misses none of the evaluated weeks, but its average pointwise width is $6{,}078.54$ cases, indicating overcoverage (see Section~\ref{sec:ablation-study} of the Supplementary Material). This distinction motivates the two-layer structure of our framework: flow-matching SBI supplies rapid approximate parameter draws, while CRC chooses among the induced simulator-output sets using prediction error on a separate synthetic calibration sample, targeting the quantity that matters downstream---predictive miscoverage---with explicit finite-sample risk control under exchangeability.

\subsection{Conformal Risk Control}
\label{subsec:CRC}

Conformal risk control (CRC) extends split conformal prediction from ensuring marginal coverage to controlling a user-specified expected loss, retaining finite-sample validity under exchangeability \citep{angelopoulos2024conformal,vovk2005algorithmic,romano2019conformalized,angelopoulos2023gentle}. Let $\mathcal{D}_{\mathrm{cal}}:=\{(X_i,Y_i)\}_{i=1}^n$ denote a separate calibration dataset and $(X_{n+1},Y_{n+1})$ a new test point, where $X_i$ represents the input features and $Y_i$ the response. Throughout, we use the term \emph{calibration} in two senses: \emph{computer model calibration} (learning simulator parameters $\theta$) and \emph{CRC calibration} (using a separate sample $\mathcal{D}_{\mathrm{cal}}$ to tune statistical risk).

Let $\Lambda\subseteq\mathbb{R}$ be a nonempty, totally ordered set of admissible threshold values whose largest element $\lambda_{\max}=\sup\Lambda$ exists, and let $\{\widehat C_\lambda\}_{\lambda\in\Lambda}$ be a family of prediction sets in which larger $\lambda$ yields more conservative sets. For each observation $i$, define a bounded loss
\begin{equation*}
L_i(\lambda) := L\!\left(\widehat C_\lambda(X_i),Y_i\right), \qquad L_i(\lambda)\in(-\infty,B],
\end{equation*}
where $B$ is a known upper bound; $L_i(\lambda)$ measures the discrepancy between $\widehat C_\lambda(X_i)$ and $Y_i$ (e.g., empirical miscoverage). The goal is a data-dependent threshold $\widehat\lambda$ such that $\mathbb{E}[L_{n+1}(\widehat\lambda)]\le \alpha$ for a target level $\alpha\in(0,1)$. Assuming each $L_i(\lambda)$ is non-increasing and right-continuous in $\lambda$ with $L_i(\lambda_{\max})\le \alpha$ almost surely, let $\widehat R_n(\lambda):=n^{-1}\sum_{i=1}^n L_i(\lambda)$ be the empirical calibration risk and select
\begin{equation}
\label{eq:crc-threshold}
\widehat\lambda = \inf\left\{\lambda\in\Lambda:\; \frac{n}{n+1}\widehat R_n(\lambda)+\frac{B}{n+1}\le\alpha\right\}
= \inf\left\{\lambda\in\Lambda:\; \widehat R_n(\lambda)\le\alpha_{\mathrm{adj}}\right\},
\end{equation}
where $\alpha_{\mathrm{adj}} = (n+1)\alpha/n - B/n$ is the finite-sample adjusted target level, with $\widehat\lambda=\lambda_{\max}$ if the set is empty. \citet{angelopoulos2024conformal} show that if $L_i(\lambda),\, i=1,\dots,n+1$, are exchangeable, then $\mathbb{E}[L_{n+1}(\widehat\lambda)] \le \alpha$. In our setting, CRC thus provides a rigorous mechanism for constructing prediction intervals with controlled expected miscoverage when nominal credible levels cannot be trusted.

\subsection{Flow Matching}
\label{subsec:flow-matching}

We approximate the target distribution of the calibration parameters using flow matching, which learns a time-dependent vector field transporting a simple reference distribution to a complex target \citep{lipman2023flowmatching}. We use the Gaussian conditional probability path
\begin{equation*}
p_t(x\mid z)=\mathcal{N}\big(x;\,tz,\,(1-(1-\sigma_\mathrm{min})t)^2 I_d\big),\qquad t\in[0,1],
\end{equation*}
where $x \in \mathbb{R}^d$ is the state variable, $z \sim p_{\mathrm{data}}$ is a target sample, and $\sigma_{\mathrm{min}} \in (0,1)$ is a small variance floor. The trajectory starts from standard Gaussian noise at $t=0$ and concentrates around $z$ as $t\to 1$, so that marginally $p_0(x)=\mathcal{N}(0,I_d)$ and $p_1(x)\approx p_{\mathrm{data}}$. A neural network $v_t^{\phi}(x)$ with weights $\phi$ is trained with the conditional flow matching (CFM) objective
\begin{equation*}
\mathcal{L}_{\mathrm{CFM}}(\phi)=\mathbb{E}_{t\sim\pi,\,z\sim p_{\mathrm{data}},\,x\sim p_t(\cdot\mid z)}\left[\big\|v_t^{\phi}(x)-v_t^{\mathrm{flow}}(x\mid z)\big\|^2\right],
\qquad
v_t^{\mathrm{flow}}(x\mid z)=\frac{z-(1-\sigma_\mathrm{min})x}{1-(1-\sigma_\mathrm{min})t},
\end{equation*}
where $\pi(t)$ is a sampling distribution over $[0,1]$ and $v_t^{\mathrm{flow}}$ is the closed-form conditional vector field generating the Gaussian path. This objective is a tractable, simulation-free surrogate for learning the target transport \citep{lipman2023flowmatching,lipman2024flowmatchingguidecode}; it is minimized by stochastic gradient descent on minibatch Monte Carlo estimates without solving an ODE during training. After training, approximate samples are generated by integrating the learned ordinary differential equation (ODE)
\begin{equation*}
X_0\sim\mathcal{N}(0,I_d),\qquad dX_t=v_t^{\phi}(X_t)\,dt,\qquad t\in[0,1],
\end{equation*}
and taking the terminal state $X_1$.

Because our goal is to sample from the \emph{conditional} posterior given observed covariates $y$ (e.g., simulator outputs), we condition the vector field on $y$ and employ classifier-free guidance (CFG) at sampling time,
\begin{equation*}
\tilde v_t(x\mid y)=\bigl(1-w_t\bigr)v_t^{\phi}(x\mid\emptyset)+w_t v_t^{\phi}(x\mid y),\qquad w_t>1,
\end{equation*}
where $\emptyset$ denotes a learnable unconditional null-embedding \citep{ho2022classifierfree,zheng2023guidedflows}: on condition-dropped training examples the raw condition $y=[A,Z]\in\mathbb{R}^{a+T}$, with $a=\dim(A)$, is replaced by a trainable vector $e_{\emptyset}\in\mathbb{R}^{a+T}$ that is optimized jointly with the network, so the conditional and unconditional fields share all other parameters. A general formulation of flow matching, including marginal and conditional paths, objective equivalence, and guided flows, is provided in Section~\ref{sec:si-flow-matching} of the Supplementary Material.

\section{Prediction Uncertainty Controlled Calibration}
\label{sec:methods}

This section develops our framework for constructing simulation-based calibration sets with finite-sample control over expected predictive miscoverage. Throughout, the observed sequence, simulator output, and discrepancy have common dimension $T$: for $\theta\in\Theta$, $Y(\theta)=(Y_1(\theta),\ldots,Y_T(\theta))^\top$, $\delta=(\delta_1,\ldots,\delta_T)^\top$, and $Z=Y(\theta)+\delta=(Z_1,\ldots,Z_T)^\top\in\mathbb{R}^T$. When the loss is evaluated only at scientifically relevant coordinates, $\mathcal{I}(Z)\subseteq\{1,\ldots,T\}$ denotes their deterministic index set and $T(Z)=|\mathcal{I}(Z)|$; otherwise $\mathcal{I}(Z)=\{1,\ldots,T\}$ and $T(Z)=T$. Let $\alpha\in(0,1)$ denote the target risk level, $n$ the size of the CRC calibration dataset, and $N$ the number of amortized posterior draws used to construct each candidate parameter set.

To tune the conformal parameter sets, we construct a synthetic CRC calibration dataset $\mathcal{D}_n:=\{(\tilde{A}_i,\tilde{Z}_i,\theta_i)\}_{i=1}^n$ that mimics the joint distribution of the parameters and the observables: parameters $\theta_i$ are drawn from the trained flow-matching posterior approximation conditioned on the target observation, and synthetic observations $\tilde{Z}_i=Y(\theta_i)+\tilde{\delta}_i$ are formed with simulated discrepancies $\tilde{\delta}_i$ drawn from a sufficiently flexible distribution, since exact knowledge of the discrepancy law is not assumed. In some problems we optionally extract auxiliary summaries $\tilde{A}_i=A(\theta_i)\in\mathcal{A}$, forward-model outputs assumed to be observed exactly. The construction of $\mathcal{D}_n$ is detailed in Section~\ref{subsec:calibration-data-generation}. Section~\ref{subsec:miscoverage-risk} presents the miscoverage loss and parameter-set constructions, Section~\ref{subsec:flow-matching-network} the conditional flow-matching architecture, and Section~\ref{subsec:discrepancy} the discrepancy-simulation strategy.

\subsection{Miscoverage Risk}
\label{subsec:miscoverage-risk}

For the $i$-th calibration sample, define $\mathcal{I}_i:=\mathcal{I}(\tilde Z_i)$ and $T_i:=|\mathcal{I}_i|$. Let $\Lambda\subseteq\mathbb{R}$ be a nonempty, totally ordered set of admissible thresholds with largest element $\lambda_{\max}=\sup\Lambda$. Denote the candidate parameter set given an auxiliary summary $A$ and observed data $Z$ by $\mathcal{C}_\lambda(A,Z)$, $\lambda\in\Lambda$, and require the family to be nested: $\lambda\le\lambda'$ implies $\mathcal{C}_\lambda(A,Z)\subseteq\mathcal{C}_{\lambda'}(A,Z)$. Let $m_{i,\lambda}:=|\mathcal{C}_\lambda(\tilde A_i,\tilde Z_i)|$ and let $N_i^{\mathrm{ref}}:= m_{i,\lambda_{\max}}\ge 1$ be the reference set size. Define the fixed lower and upper tail ranks by
\begin{equation*}
k_i^- := \left\lceil N_i^{\mathrm{ref}}\frac{\rho}{2}\right\rceil,\qquad
q_i^+ := \max\left\{1,\left\lfloor N_i^{\mathrm{ref}}\left(1-\frac{\rho}{2}\right) \right\rfloor\right\},\qquad
k_i^+ := N_i^{\mathrm{ref}}-q_i^+ +1,
\end{equation*}
where $\rho\in(0,1)$ is a user-specified interval-tail hyperparameter; $k_i^-$ is the rank from the lower tail and $k_i^+$ the rank from the upper tail. These ranks are computed once from $N_i^{\mathrm{ref}}$ and reused for every candidate $\lambda$.

For any finite parameter set $\Theta$ and coordinate $u$, let $Y_{u,(r)}(\Theta)$, $1\le r\le |\Theta|$, denote the $r$-th order statistic of $\{Y_u(\theta):\theta\in\Theta\}$. For a candidate set with enough retained draws, i.e., $m_{i,\lambda}\ge k_i^-+k_i^+-1$, define the fixed-rank prediction interval at coordinate $u$ by $[b_{iu}^-(\lambda), b_{iu}^+(\lambda)]$, where $b_{iu}^-(\lambda):=Y_{u,(k_i^-)}(\mathcal{C}_\lambda(\tilde A_i,\tilde Z_i))$, $r_{i,\lambda}^+:=m_{i,\lambda}-k_i^+ +1$, and $b_{iu}^+(\lambda):= Y_{u,(r_{i,\lambda}^+)}(\mathcal{C}_\lambda(\tilde A_i,\tilde Z_i))$. The fixed-rank empirical miscoverage loss is then defined as follows. If $T_i=0$, set $L_i(\lambda)=0$. If $T_i>0$ but $m_{i,\lambda} < k_i^-+k_i^+-1$, set $L_i(\lambda)=1$. Otherwise,
\begin{equation}
L_i(\lambda)
:=
\frac{1}{T_i}\sum_{u\in\mathcal{I}_i}
\mathbb{I}\!\left\{
\tilde Z_{iu}\notin
\left[b_{iu}^-(\lambda),b_{iu}^+(\lambda)\right]
\right\}.
\label{eq:empirical-miscoverage-loss}
\end{equation}
The fixed-rank construction guarantees monotonicity. Because the candidate sets are nested in $\lambda$, adding draws can only decrease the $k_i^-$-th smallest and increase the $k_i^+$-th largest simulated value, so $b_{iu}^-(\lambda)$ is non-increasing and $b_{iu}^+(\lambda)$ non-decreasing in $\lambda$ and the intervals expand monotonically with $\lambda$. Consequently the threshold selected by CRC in equation~\eqref{eq:crc-threshold} favors the narrowest bands that satisfy the empirical constraint.

We introduce two formulations for the candidate sets $\mathcal{C}_\lambda(A,Z)$. For each conditioning pair $(A,Z)$, consider the conditional ODE of Section~\ref{subsec:flow-matching}, $dX_t = v_t^\phi(X_t \mid A,Z)dt$, $X_0 = x_0$, and let $\psi_{0,t}^{\phi}(x_0 \mid A,Z) := X_t$ denote the induced flow map, with set-valued pushforward $\psi_{0,t}^{\phi}(U \mid A,Z) := \{\psi_{0,t}^{\phi}(u \mid A,Z) : u \in U\}$ for finite $U \subset \mathbb{R}^d$. In both formulations, increasing $\lambda$ expands the set, driving down the fixed-rank miscoverage loss.

(i) \emph{Radius-based set.} This approach truncates the base distribution prior to the flow transformation. Let $S_N=\{\xi_j\}_{j=1}^N$ be an ensemble of $N$ base samples $\xi_j\stackrel{iid}{\sim}\mathcal{N}(0,I_d)$ and $B_\lambda(0)=\{x\in\mathbb{R}^d:\|x\|_2\le\lambda\}$ the $L_2$-ball of radius $\lambda$. We define
\begin{equation*}
\mathcal{C}_\lambda(A,Z):=\psi_{0,1}^{\phi}\big(B_\lambda(0)\cap S_N\mid A,Z\big),
\end{equation*}
with admissible threshold domain $\Lambda_{\mathrm{rad}}=[0,\max_{1\le j\le N}\|\xi_j\|_2]$, whose largest element retains the complete base ensemble.

(ii) \emph{Posterior-based set.} Alternatively, we threshold the conditional posterior density. Let $\Theta_N^\phi(A,Z):=\psi_{0,1}^{\phi}(S_N\mid A,Z)$ be the generated parameter ensemble and, for any finite set $\Theta$, let $Q_\rho\{f(\theta):\theta\in\Theta\}$ denote the empirical $\rho$-quantile of $f$ over $\Theta$. The posterior-based set retains the highest-density draws up to $\lambda \in [0,1]$:
\begin{equation*}
\mathcal{C}_\lambda(A,Z):=\left\{\theta\in\Theta_N^\phi(A,Z):\log p_1^\phi(\theta\mid A,Z)\ge Q_{1-\lambda}\!\left\{\log p_1^\phi(\vartheta\mid A,Z):\vartheta\in\Theta_N^\phi(A,Z)\right\}\right\},
\end{equation*}
with $\Lambda_{\mathrm{pos}}=[0,1]$; in practice, $\mathcal{C}_\lambda(A,Z)$ consists of the top $\lceil\lambda N\rceil$ generated draws ranked by conditional log-density. The conditional log-density is computed along the probability flow ODE via the continuous normalizing-flow (CNF) identity \citep{chen2018neuralode,lipman2023flowmatching}
\begin{equation}
\label{eq:CNF}
\frac{d}{dt}\log p_t^\phi(X_t\mid A,Z)=-\operatorname{div}v_t^\phi(X_t\mid A,Z),
\end{equation}
where $X_t$ traces the trajectory from $X_0 \sim p_0^\phi(\theta\mid A_{\mathrm{obs}},Z_{\mathrm{obs}})=\mathcal{N}(0, I_d)$ to $X_1 = \theta\sim p_1^\phi(\theta\mid A_{\mathrm{obs}},Z_{\mathrm{obs}})$.

\subsection{Selection of the trimming parameter.}
\label{subsec:rho-selection}
After fitting the flow, the tail hyperparameter $\rho$ is selected separately for each construction $j\in\{\mathrm{rad},\mathrm{pos}\}$ using simulated tuning data generated by the mechanism of Section~\ref{subsec:calibration-data-generation}, independently of the final CRC calibration sample: a tuning CRC sample of size $n_{\mathrm{tune}}=100$ and a validation sample of size $n_v=500$. For each $\rho$ on the grid $\mathscr{G}=\{0.01,0.02,\ldots,0.20\}$, we compute a tuning threshold $\widehat\lambda_j^{\mathrm{tune}}(\rho)$ from equation~\eqref{eq:crc-threshold} with $n_{\mathrm{tune}}$ in place of $n$, and evaluate it on the validation sample using two criteria: the validation miscoverage and the mean pointwise interval length,
\begin{align}
\widehat R_{j,\mathrm{val}}(\rho)
&=\frac{1}{n_v}\sum_{v=1}^{n_v}
L_{jv}^{\mathrm{val}}\!\left(\widehat\lambda_j^{\mathrm{tune}}(\rho);\rho\right),\nonumber\\
W_j(\rho)
&=\frac{1}{n_vT}\sum_{v=1}^{n_v}\sum_{u=1}^{T}
\left[b_{jvu}^{+}\!\left(\widehat\lambda_j^{\mathrm{tune}}(\rho);\rho\right)
-b_{jvu}^{-}\!\left(\widehat\lambda_j^{\mathrm{tune}}(\rho);\rho\right)\right],
\label{eq:rho-validation}
\end{align}
where $L_{jv}^{\mathrm{val}}$ is the loss in equation~\eqref{eq:empirical-miscoverage-loss} for validation observation $v$ and $b_{jvu}^{\pm}$ are the corresponding interval endpoints at coordinate $u$; we set $W_j(\rho)=+\infty$ if any validation interval lacks enough retained draws to be defined. The first criterion ensures that the tuned procedure attains the target risk on fresh data, and the second measures sharpness. We select the admissible value with the sharpest intervals,
\begin{equation}
\widehat\rho_j\in
\operatorname*{arg\,min}_{\rho\in\mathscr G_j^{\mathrm{adm}}}
W_j(\rho),
\qquad
\mathscr G_j^{\mathrm{adm}}=
\{\rho\in\mathscr G:\ \widehat R_{j,\mathrm{val}}(\rho)\le\alpha,\ W_j(\rho)<\infty\},
\label{eq:rho-selection}
\end{equation}
and the rule yields no selection if $\mathscr G_j^{\mathrm{adm}}$ is empty. We then freeze $\rho_j=\widehat\rho_j$, generate a fresh final CRC calibration sample, and recompute the final threshold $\widehat\lambda_j$. Conditional on the fitted model and tuning information, the CRC bound retains its form provided the assumptions of Section~\ref{subsec:CRC} hold for the final calibration and target losses \citep{angelopoulos2024conformal}.

\subsection{Deep Approximate Distribution with Flow Matching}
\label{subsec:flow-matching-network}

We parameterize the conditional vector field $v_t^\phi(x\mid A,Z)$ by a deep neural network with inputs $(x,A,Z,t)$, where $x\in\mathbb{R}^d$ is the latent state at time $t\in[0,1]$ \citep{dax2023flowmatching,lipman2023flowmatching,lipman2024flowmatchingguidecode}. The conditioning information $(t,A,Z)$ enters through a context vector $c_t=[h_t,\,h_Z,\,A]$. The time embedding $h_t\in\mathbb{R}^{256}$ is obtained by passing random Fourier features of $t$ \citep{rahimi2007randomfeatures} through a two-layer SiLU network. The sequence representation $h_Z$ is built by dividing $Z\in\mathbb{R}^T$ into overlapping patches of width $7$ and stride $2$, linearly embedding each patch into $\mathbb{R}^{128}$, prepending a learned start-of-sequence token, adding positional encodings, and passing the result through a four-layer Transformer encoder \citep{vaswani2017attention}. The encoder output is aggregated by learned-query multihead-attention pooling with $q=4$ queries and $4$ heads \citep{lee2019settransformer}, projected back to $\mathbb{R}^{128}$, and augmented with two geometric features, $\log\{1+\max_u Z_u\}$ and the normalized location of the maximum, giving $h_Z\in\mathbb{R}^{130}$.

Given $x$ and $c_t$, the backbone consists of $N_{\mathrm{blk}}=4$ fixed-width residual blocks of hidden dimension $256$ with Adaptive Layer Normalization (AdaLN) and Swish-Gated Linear Units (SwiGLU) \citep{shazeer2020glu,peebles2023scalable}: in each block, the context vector produces a scale and shift applied to the layer-normalized state, which is then passed through a gated linear unit with dropout and added residually. The final normalized state is concatenated with $c_t$ and passed through $d$ coordinate-specific output heads to produce $v_t^\phi(x\mid A,Z)\in\mathbb{R}^d$. The context projections, residual output projections, and output heads are initialized to zero, so each block initially acts as the identity and the initial vector field is zero, which improves numerical stability early in training. The complete layer-by-layer specification is provided in Section~\ref{sec:si-architecture} of the Supplementary Material.

\subsection{Handling Data-Model Discrepancy}
\label{subsec:discrepancy}

Data-model discrepancy---a mismatch between the model output $Y(\theta)$ and the observed data $Z$ even at the best input setting---is a central challenge in computer model calibration, and ignoring it can bias inference on $\theta$ \citep{bhatnagar2022computer,brynjarsdottir2014learning,tuo2015efficient}. To make the flow-matching model robust to discrepancy, we train it on model outputs contaminated with synthetic discrepancy terms sampled from Gaussian processes with varying hyperparameters. Exposure to a diverse array of discrepancies encourages the learned inverse map to identify features that are informative for recovering the physical parameters while accounting for discrepancy.

\subsubsection{Training Data Generation}
\label{subsec:training-data-generation}

We first draw a pool of $N_{\text{param}}$ parameter configurations $\theta_k$, $k \in \{1, \dots, N_{\text{param}}\}$, via Latin Hypercube Sampling (LHS) over the prior bounds \citep{mckay1979lhs} and evaluate the simulator to obtain the clean trajectories $Y(\theta_k)$ and auxiliary outputs $\tilde{A}_k = A(\theta_k)$. Separately, we generate a pool of $N_{\text{noise}}$ raw discrepancy realizations $\tilde{\delta}_{j}$, $j \in \{1, \dots, N_{\text{noise}}\}$, from a zero-mean Gaussian process (GP) \citep{bhatnagar2022computer} whose marginal variance scales with the temporal variability of the clean output. Writing $\sigma_Y^2(\boldsymbol{\theta}) = \frac{1}{T} \sum_{u=1}^T ( Y_u(\boldsymbol{\theta}) - \bar{Y} (\boldsymbol{\theta}) )^2$ with $\bar{Y} (\boldsymbol{\theta})=\frac{1}{T}  \sum_{u=1}^T  Y_u(\boldsymbol{\theta})$, the covariance between indices $u_1, u_2 \in \{1, \ldots, T\}$ is
\begin{equation}
K(u_1, u_2;\, \boldsymbol{\theta})
  = \sigma_Y^2(\boldsymbol{\theta})\left[
      \zeta\,\mathbb{I}(u_1 = u_2)
    + \kappa \exp\!\left( -\frac{|u_1 - u_2|}{\tau} \right)
  \right].
\label{eq:gp-discrepancy-kernel}
\end{equation}
The prefactor $\sigma_Y^2(\boldsymbol{\theta})$ keeps the discrepancy magnitude commensurate with the amplitude of each trajectory, so that the nugget $\zeta$, partial sill $\kappa$, and range $\tau$ control the noise-to-signal ratio, the relative strength of the temporally correlated component, and the correlation length, respectively. These hyperparameters are sampled via LHS from user-specified bounds for each realization. The training set pairs every parameter configuration $k$ with every noise realization $j$,
\begin{equation}
\tilde{Z}_{k,j, u} = Y_u(\theta_k) + \tilde{\delta}_{j, u},
\label{eq:training-contaminated-sequence}
\end{equation}
with contaminated states clipped to their theoretical domain (e.g., nonnegative counts), yielding $\mathcal{D}_{\text{train}} = \{(\tilde{A}_k, \tilde{Z}_{k,j}, \theta_k)\}$ of size $N_{\text{train}} = N_{\text{param}} \times N_{\text{noise}}$.

\subsubsection{Calibration Data Generation for Conformal Risk Control}
\label{subsec:calibration-data-generation}

After training, the $n$ CRC calibration samples in $\mathcal{D}_n$ are generated independently, conditional on the fitted flow and the observed conditioning pair. We draw $\theta_i\sim p_1^\phi(\theta\mid A_{\mathrm{obs}},Z_{\mathrm{obs}})$ via ODE integration, each requiring only a forward pass through the trained flow rather than a new posterior computation, clamp the draws to their physical bounds, and pass each through the simulator to obtain $Y(\theta_i)$ and $\tilde A_i$. Rather than reusing the training noise pool, we independently sample new GP hyperparameters and generate a unique raw GP draw $\tilde{\delta}_i$ for each $\theta_i$, and then apply a nonnegative, problem-specific state-dependent envelope $M_u\{Y(\theta_i)\}$ to reproduce the heteroscedastic scale pattern of the target observation:
\begin{equation}
\tilde Z_{i,u}=Y_u(\theta_i)+M_u\{Y(\theta_i)\}\tilde{\delta}_{i,u},\qquad u=1,\ldots,T.
\label{eq:crc-calibration-observation}
\end{equation}
The scale at coordinate $u$ may depend on the entire clean trajectory; both studies below use
\begin{equation}
\label{eq:envelope}
M_u\{Y(\theta)\}=\left\{\frac{Y_u(\theta)-\min\limits_{1\le s\le T}Y_s(\theta)}{\max\limits_{1\le s\le T}Y_s(\theta)-\min\limits_{1\le s\le T}Y_s(\theta)}\right\}^{1/2},
\end{equation}
so that discrepancies are attenuated in low-signal regions. After modulation, the sequence is clipped to its physical bounds. The synthetic calibration samples are therefore independent draws from
\begin{equation}
\label{eq:joint-distribution}
P_{\mathrm{cal}}(A,Z,\theta)=p_{\delta,M}\{Z\mid Y(\theta)\}\,\mathbb{I}\{A=A(\theta)\}\,p(\theta),
\end{equation}
where $p(\theta)=p_1^\phi(\theta\mid A_{\mathrm{obs}},Z_{\mathrm{obs}})$ and $p_{\delta,M}$ is the conditional law induced jointly by the raw GP, the envelope $M$, and clipping.

The two discrepancy mechanisms serve different purposes. Unmodulated discrepancies in $\mathcal D_{\mathrm{train}}$ are a data-augmentation device that makes the learned inverse map robust to severe discrepancies across the full trajectory, whereas the envelope in equation~\eqref{eq:crc-calibration-observation} makes the synthetic calibration data reproduce the state-dependent variability anticipated in the observed data, encouraging exchangeability of the target observation with the calibration data. The finite-sample CRC conclusion rests on the working assumption that the target observation $(A_{\mathrm{obs}},Z_{\mathrm{obs}})$ and the $n$ synthetic observations $\{(\tilde{A}_i, \tilde{Z}_i)\}_{i=1}^n$ are jointly exchangeable under this envelope-modulated mechanism. Since the loss $L_i(\lambda)$ depends only on the observations and not on $\theta_i$, the CRC result of Section~\ref{subsec:CRC} applies under this assumption \citep{angelopoulos2024conformal}.

\section{Simulation Study}
\label{sec:simulations}

We evaluate the proposed framework using the synthetic computer model and observations of \citet{bhatnagar2022computer}. Unlike the application in Section~\ref{sec:applications}, the synthetic output $Y(\theta)$ is available in closed form without a numerical solver, providing an inexpensive testbed. Throughout this section, the target miscoverage level is $\alpha=0.1$.

\subsection{Synthetic Model Outputs and Observational Data}
\label{subsec:synthetic-data}

\begin{figure}[!tbp]
\centering
\includegraphics[width=0.62\textwidth]{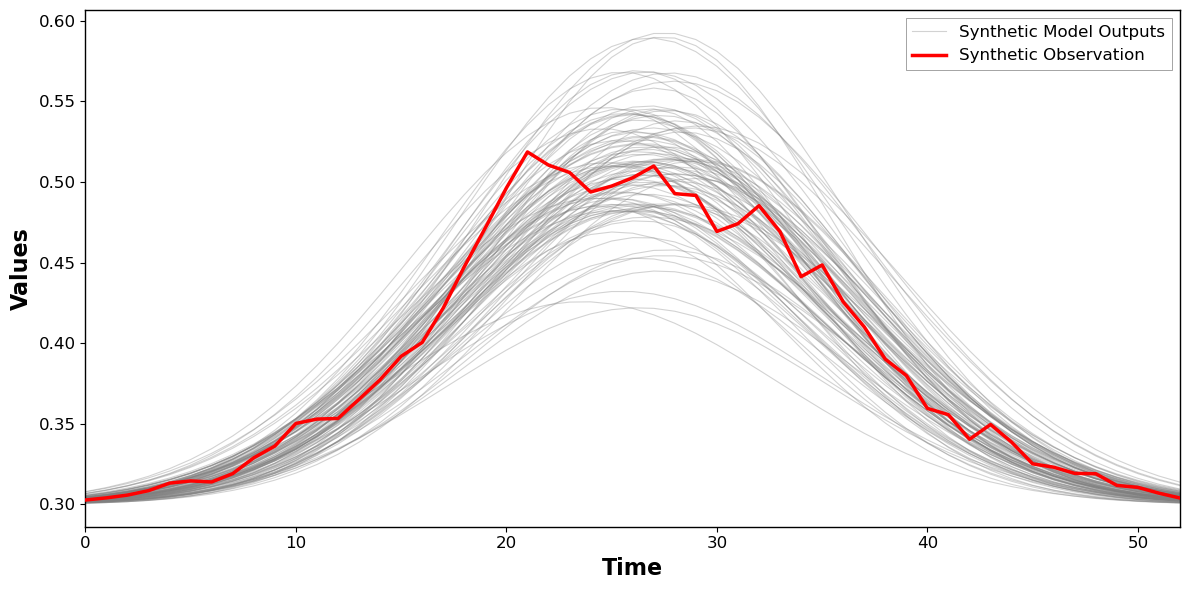}
\caption{Synthetic observation and representative model outputs from the simulation-study forward model.}
\label{fig:simulation-example-model-outputs}
\end{figure}

The deterministic forward model produces smooth unimodal curves governed by $\boldsymbol{\theta} = (\theta_1,\theta_2,\theta_3)^\top \in [0,1]^3$. For $T=53$ time points $s(u)=\frac{4(u-1)}{T-1}-2$ equally spaced on $[-2,2]$, the output at index $u$ is
\begin{equation*}
Y_u(\boldsymbol{\theta}) = 0.3 + \frac{\theta_1 + 0.3}{\sqrt{2\pi(\theta_3 + 0.1)}} \exp\!\left[-\frac{(s(u) - \theta_2 + 0.5)^2}{\theta_3 + 0.1}\right],
\end{equation*}
where $\theta_1$ controls the amplitude, $\theta_2$ the peak location, and $\theta_3$ the dispersion. Representative outputs and the synthetic observation are shown in Figure~\ref{fig:simulation-example-model-outputs}.

All datasets share the GP discrepancy family of Section~\ref{subsec:discrepancy}, with the kernel in equation~\eqref{eq:gp-discrepancy-kernel} and hyperparameters sampled via LHS from range proportion $\tau_{\mathrm{prop}}\in[0.05,0.20]$ (so that $\tau=\tau_{\mathrm{prop}}\cdot T$), partial sill $\kappa\in[0.05^2,0.25^2]$, and nugget $\zeta\in[0.005^2,0.025^2]$.

\emph{Training data.} We draw $N_{\mathrm{train}}=5{,}000$ parameter configurations via LHS over $[0,1]^3$ and $N_{\mathrm{noise}}=30$ GP hyperparameter vectors, and pair every configuration with every noise realization via equation~\eqref{eq:training-contaminated-sequence}, yielding $150{,}000$ contaminated trajectories. As discussed in Section~\ref{subsec:discrepancy}, the training discrepancies are left unscaled by any state-dependent envelope so that the network sees severe discrepancies at all signal levels.

\emph{Synthetic observation data.} We fix a ground-truth parameter $\boldsymbol{\theta}^*$ and draw a raw GP discrepancy $\varepsilon_{\mathrm{obs}}$ with covariance given by equation~\eqref{eq:gp-discrepancy-kernel} at $\boldsymbol{\theta}^*$, modulated by the envelope of equation~\eqref{eq:envelope}:
\begin{equation}
Z_{\mathrm{obs},u}=Y_u(\boldsymbol{\theta}^*)+M_u\{Y(\boldsymbol{\theta}^*)\}\varepsilon_{\mathrm{obs},u},\qquad u=1,\ldots,T.
\label{eq:simulation-observation}
\end{equation}

\emph{Calibration data.} A separate CRC calibration dataset $\mathcal{D}_n$ of size $n=100$ is built following Section~\ref{subsec:calibration-data-generation}: parameters are sampled through the trained flow map $\psi_{0,1}^\phi(\cdot \mid A_{\mathrm{obs}}, Z_{\mathrm{obs}})$ and clamped to $[0,1]^3$, and each draw receives a unique envelope-modulated GP discrepancy as in equation~\eqref{eq:crc-calibration-observation}. The synthetic target and the CRC calibration observations thus deliberately share the same discrepancy mechanism (GP family, hyperparameter law, and envelope), isolating performance under the nominal specification; matching the discrepancy mechanisms does not by itself establish joint exchangeability, which remains a working assumption (Section~\ref{subsec:calibration-data-generation}). The thresholds $\widehat{\lambda}_{\mathrm{rad}}$ and $\widehat{\lambda}_{\mathrm{pos}}$ are calibrated once on $\mathcal{D}_n$ and held fixed thereafter.

\subsection{Implementation and Evaluation Protocol}
\label{subsec:toy-implementation-details}

We use the conditional vector-field architecture of Section~\ref{subsec:flow-matching-network} with $\sigma_\mathrm{min}=0.1$. Posterior samples are generated by integrating the conditional ODE with the Euler scheme over $100$ equally spaced steps on $[0,1]$, using classifier-free guidance scale $w=1.2$ with an interval-based scheduler \citep{ho2022classifierfree,zheng2023guidedflows,kynkaanniemi2024limitedinterval}, and the generated parameters are mapped back to the original scale and clamped to $[0,1]^3$. For both set constructions we draw $N=100$ latent base samples $\xi_j \stackrel{\mathrm{iid}}{\sim} \mathcal{N}(0, I_d)$ and transport them through the flow; the same latent pool is shared across all $n$ calibration samples so that differences among parameter sets are driven by the conditioning observations rather than base-space sampling noise.

The miscoverage loss of equation~\eqref{eq:empirical-miscoverage-loss} uses the independently selected values $\rho_{\mathrm{rad}}=\rho_{\mathrm{pos}}=0.03$ (Section~\ref{subsec:rho-selection}) and the relevant-coordinate set $\mathcal I(Z)=\{u:Z_u\ge\eta\}$ with $\eta=0.301$. Since the clean simulator has baseline $0.3$ and minimum attainable peak $0.3+0.3/\sqrt{2\pi(1.1)}\approx0.414$ over $[0,1]^3$, $\eta$ lies only $0.001$ (about $0.88\%$ of the minimum dynamic range) above the baseline and excludes only essentially flat background coordinates. For the radius-based sets, the ODE trajectories of all $N$ latent draws conditioned on each calibration observation are precomputed together with the base-space norms $\|\xi_j\|_2$, reducing the threshold search to a filtering operation; for the posterior-based sets, the conditional log-densities computed via equation~\eqref{eq:CNF} are cached alongside the generated parameters. In both cases $\widehat\lambda$ is obtained by applying Brent's root-finding algorithm \citep{brent1973algorithms} to $\bar{L}(\lambda)-\alpha_{\mathrm{adj}}$, where $\bar{L}(\lambda)=n^{-1}\sum_{i=1}^n L_i(\lambda)$ and $\alpha_{\mathrm{adj}}=(n+1)\alpha/n-1/n$; the search is carried out over $\lambda\in\bigl(0,\max_{1\le j\le N}\|\xi_j\|_2\bigr]$ for the radius-based sets and over $\lambda\in(0.001,0.999)$ for the posterior-based sets, with tolerance $10^{-4}$.

We use repeated held-out simulation to assess how each procedure behaves for repeated measurements of the same system. Holding $\boldsymbol{\theta}^*$ fixed, we conduct $N_{\mathrm{exp}}=200$ experiments, regenerating an independent envelope-modulated GP discrepancy with newly sampled hyperparameters according to equation~\eqref{eq:simulation-observation} in each. The CRC thresholds are estimated once from $\mathcal D_n$ and frozen before the held-out experiments, mimicking deployment to future observations. For each held-out observation we construct parameter sets of maximum size $N^{\text{ref}}=100$ and the associated predictive bands, and record (i) empirical miscoverage, (ii) average pointwise interval length (sharpness), and (iii) RMSE between the held-out observation and the band median (center accuracy). For comparison, a KOH--MCMC baseline is rerun separately for every held-out observation and its pointwise $(1-\alpha)$ posterior band is constructed, reproducing the observation-specific, non-amortized Bayesian workflow (MCMC specifications and convergence diagnostics are given in Section~\ref{sec:si-mcmc-simulation} of the Supplementary Material).

\subsection{Results}
\label{subsec:simulation-results}

\begin{figure}[p]
\centering
\begin{subfigure}{0.72\textwidth}
\centering
\includegraphics[width=\linewidth]{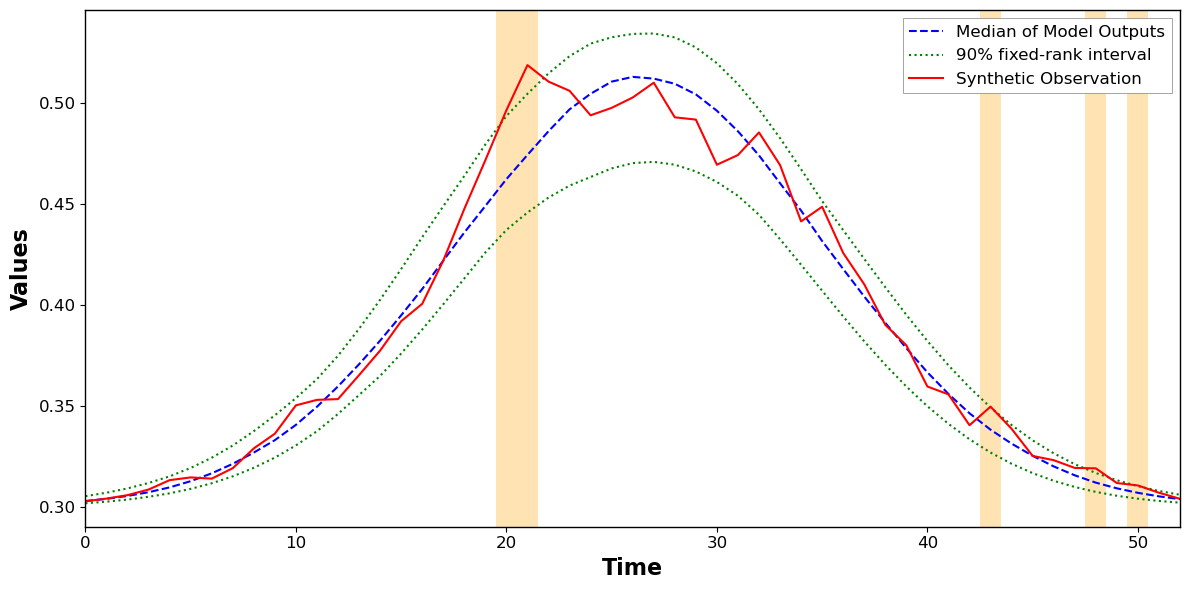}
\caption{Radius-based construction: miscoverage $0.0943$, RMSE $0.0119$, average pointwise interval length $0.0337$.}
\label{fig:sim-radius}
\end{subfigure}

\smallskip
\begin{subfigure}{0.72\textwidth}
\centering
\includegraphics[width=\linewidth]{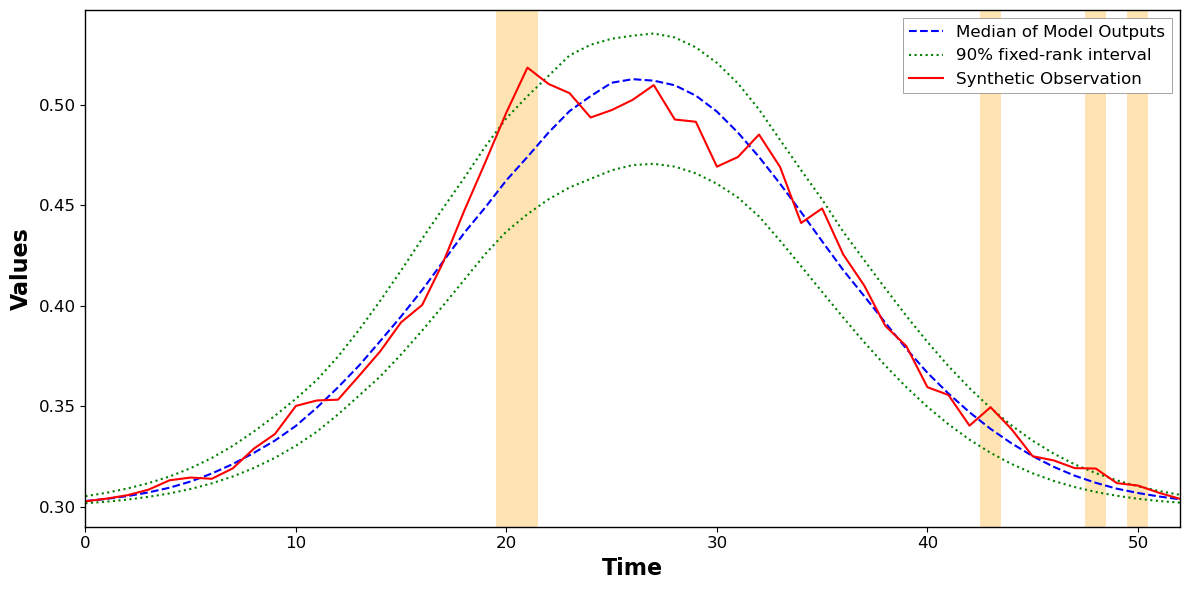}
\caption{Posterior-based construction: miscoverage $0.0943$, RMSE $0.0118$, average pointwise interval length $0.0339$.}
\label{fig:sim-likelihood}
\end{subfigure}

\smallskip
\begin{subfigure}{0.72\textwidth}
\centering
\includegraphics[width=\linewidth]{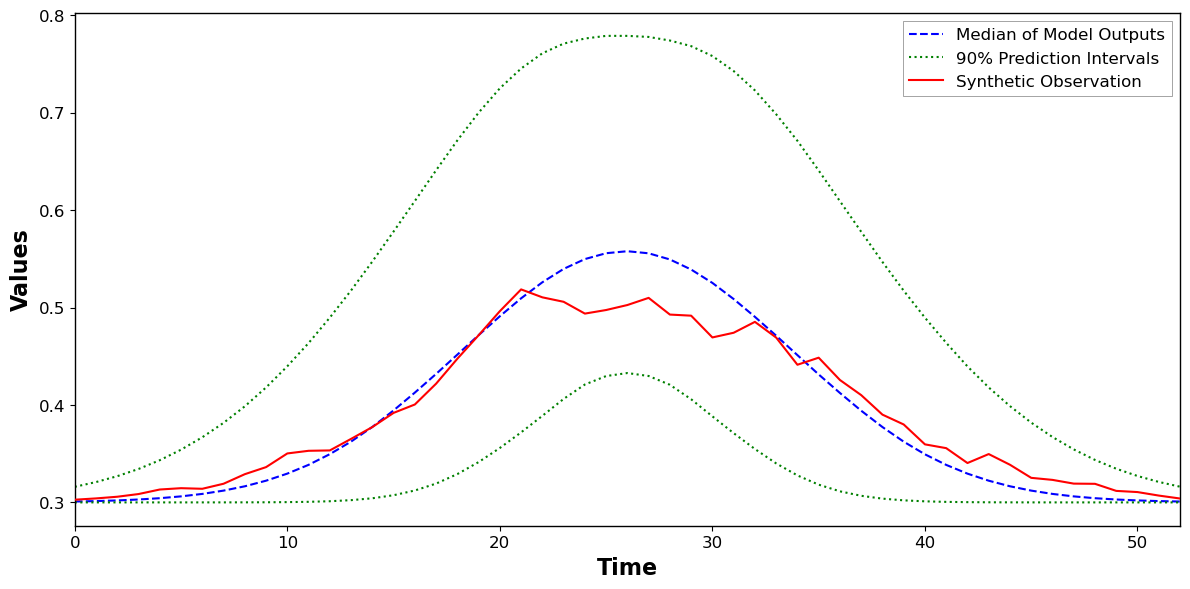}
\caption{MCMC baseline: miscoverage $0.0000$, RMSE $0.0232$, average pointwise interval length $0.2058$.}
\label{fig:sim-mcmc}
\end{subfigure}
\caption{Prediction bands for the same representative synthetic observation under the two CRC constructions and the KOH--MCMC baseline, with target miscoverage $\alpha=0.1$. Line and shading conventions are described in the text.}
\label{fig:sim-crc-intervals}
\end{figure}

Figure~\ref{fig:sim-crc-intervals} displays the predictive bands produced by the radius-based, posterior-based, and MCMC methods for a single representative synthetic observation. Here, we intentionally show a case where the radius-based and posterior-based methods lead to almost the same result. The blue dashed curve is the median of the simulator outputs over the parameter set, the green dotted curves trace the pointwise $90\%$ prediction interval, the red solid curve is the synthetic observation, and orange shading marks time points where the observation falls outside the interval (and exceeds $\eta$). On this draw, the MCMC band envelops the observation entirely at the cost of roughly $7\times$ longer pointwise intervals and roughly twice the RMSE between the predictive median and the observation, whereas the two CRC bands enclose the observation much more tightly.

\begin{figure}[!tbp]
\centering
\begin{subfigure}{0.32\textwidth}
\centering
\includegraphics[width=\textwidth]{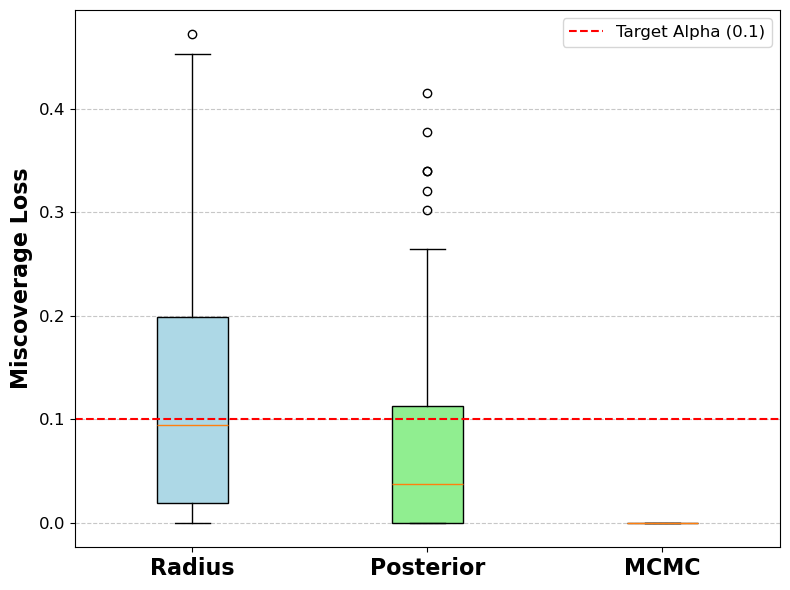}
\caption{Miscoverage}
\label{fig:sim-box-miscoverage}
\end{subfigure}\hfill
\begin{subfigure}{0.32\textwidth}
\centering
\includegraphics[width=\textwidth]{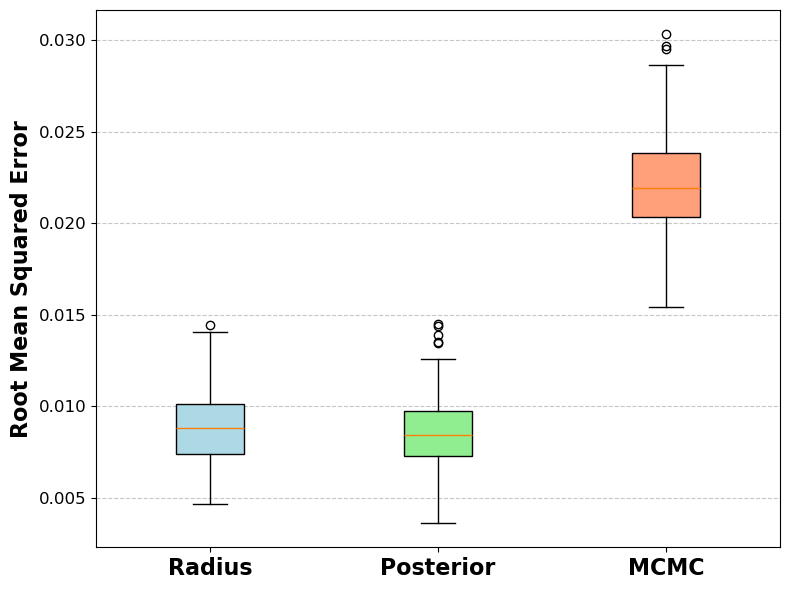}
\caption{RMSE}
\label{fig:sim-box-rmse}
\end{subfigure}\hfill
\begin{subfigure}{0.32\textwidth}
\centering
\includegraphics[width=\textwidth]{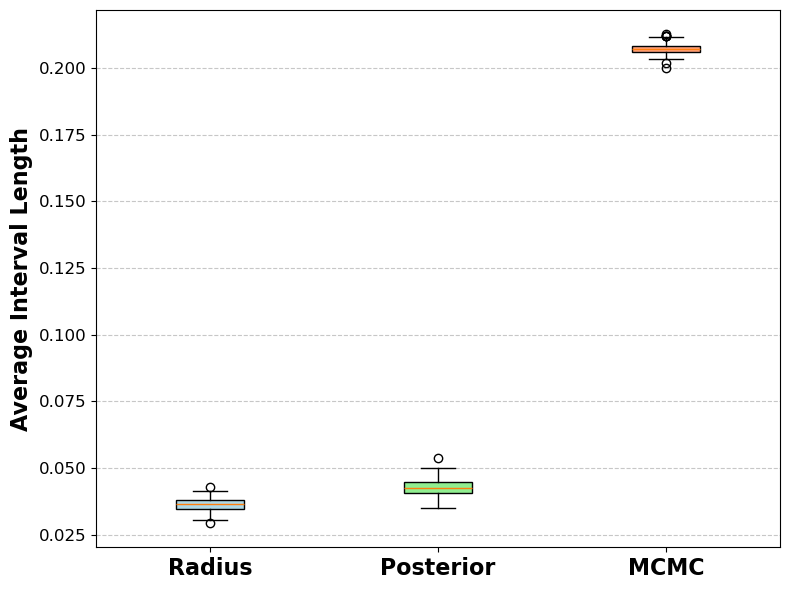}
\caption{Average interval length}
\label{fig:sim-box-length}
\end{subfigure}
\caption{Distributions of evaluation metrics across $200$ independent experiments: empirical miscoverage, RMSE between the predictive median and the synthetic observation, and average pointwise interval length. The horizontal dashed line in panel (a) marks the target level $\alpha = 0.1$.}
\label{fig:sim-box-metrics}
\end{figure}

Figure~\ref{fig:sim-box-metrics} summarizes the three metrics across the $N_{\mathrm{exp}} = 200$ experiments. The radius-based set, calibrated at $\hat\lambda_{\mathrm{rad}} = 1.3078$, attains a mean miscoverage of $0.1222$, close to the target, with some dispersion ($75\%$ quantile $0.1989$). The posterior-based set, calibrated to retain the top $\hat\lambda_{\mathrm{pos}} = 46.47\%$ of draws by conditional log-density, is more conservative with mean miscoverage $0.0721$ ($75\%$ quantile $0.1132$). The MCMC baseline attains zero miscoverage in all $200$ experiments, reflecting bands wide enough to envelop almost every realization regardless of how the discrepancy is distributed. The RMSE and interval-length panels make the cost of this overcoverage explicit: the CRC methods achieve mean RMSE $0.0089$ (radius) and $0.0085$ (posterior), roughly a third of the MCMC mean of $0.0222$, and mean pointwise interval lengths of $0.0363$ and $0.0426$, about $20\%$ of the MCMC mean of $0.2073$---a pattern consistent with parameter--discrepancy nonidentifiability widening posterior uncertainty \citep{brynjarsdottir2014learning}.

These results illustrate the central trade-off targeted by the framework. The MCMC bands reflect posterior uncertainty under an assumed Gaussian observation model but are not calibrated to a prespecified predictive-risk target \citep{syring2019calibrating}; they are over-wide on average and cannot adapt to the realized discrepancy. The CRC methods calibrate the interval width directly against the empirical miscoverage on synthetic data generated under the same flexible GP discrepancy family \citep{angelopoulos2024conformal}, producing intervals that are tight, adaptive, and near the nominal target. An ablation that bypasses the CRC step and instead retains a fixed $90\%$ of the generated draws under either selection rule (Section~\ref{sec:ablation-study} of the Supplementary Material) indicates that this behavior is driven by the CRC calibration step rather than by the set constructions themselves: for the representative observation of Figure~\ref{fig:sim-crc-intervals}, both fixed-retention rules attain zero miscoverage but produce average interval lengths of $0.0597$, about $1.8$ times those of the corresponding CRC bands, indicating overcoverage.

\section{Application to a Measles SEIR Model}
\label{sec:applications}

We now apply the framework to a real-data calibration problem: inferring the parameters of an age-structured Susceptible--Exposed--Infected--Recovered (SEIR) model from the 2010 measles outbreak in Malawi \citep{minetti2013lessons}. This application is more challenging than the simulation study in three respects. First, the forward model is a system of $84$ coupled ODEs whose evaluation is far more expensive than the closed-form curve. Second, the posterior is non-identifiable along certain directions in parameter space, complicating both inference and discrepancy modeling. Third, the data are real, so the true data-generating mechanism is unknown and the discrepancy assumption underlying the exchangeability argument can only be assessed indirectly. Our goal is to demonstrate finite-sample miscoverage control under these realistic conditions while remaining computationally favorable relative to the Bayesian baseline.

\subsection{Measles Outbreak Data and Age-Structured SEIR Model}

\begin{figure}[!tbp]
\centering
\includegraphics[width=0.62\textwidth]{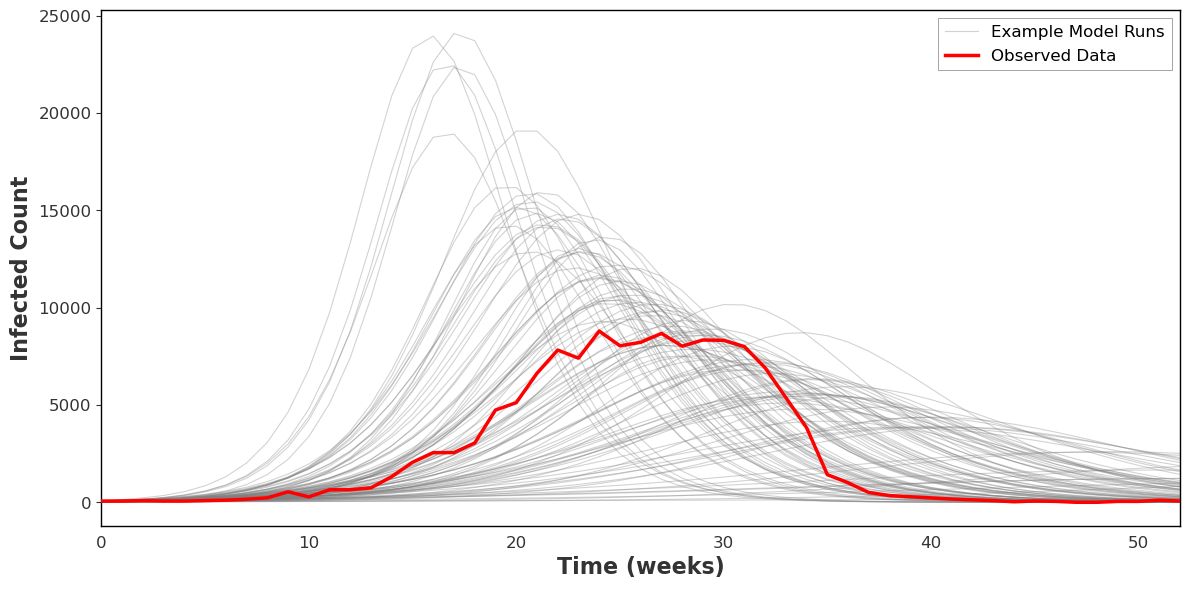}
\caption{Observed measles data and representative SEIR model outputs.}
\label{fig:real-example-model-outputs}
\end{figure}

The data, originally analyzed by \citet{minetti2013lessons} and shown in Figure~\ref{fig:real-example-model-outputs}, consist of reported weekly measles case counts $Z_{\mathrm{obs}}$ at $T=53$ epochs spanning $52$ weeks ($364$ days) and the empirical proportions of cases across three coarse age groups (0.5--5, 5--15, and 15+ years), $A_{\mathrm{obs}}=(0.42, 0.30, 0.28) \in \Delta^{2}\subset\mathbb{R}^3$, where $\Delta^{2}$ is the 2-simplex. The time series $Z_{\mathrm{obs}}$ is treated as the noisy observation subject to discrepancy, while $A_{\mathrm{obs}}$ plays the role of the auxiliary summary $A$, assumed to be observed exactly.

We adopt an age-structured SEIR formulation \citep{shea2014adaptive} in which the host population is partitioned into $K = 21$ age classes (a six-month band for 0.5--1 year, one-year bands up to 19--20 years, and a single class for 20+ years). For each class $i$, let $S_i, E_i, I_i, R_i$ denote the number of susceptible, exposed, infectious, and recovered individuals. The dynamics are
\begin{equation}
\label{eq:seir-ode}
\begin{aligned}
\frac{dS_i}{dt} &= -\lambda_i(t)\, S_i, &
\frac{dE_i}{dt} &= \lambda_i(t)\, S_i - \sigma\, E_i,\\
\frac{dI_i}{dt} &= \sigma\, E_i - \gamma\, I_i, &
\frac{dR_i}{dt} &= \gamma\, I_i,
\end{aligned}
\end{equation}
with latent and infectious periods both fixed at seven days, $\sigma = \gamma = 1/7$ \citep{shea2014adaptive}. The age-specific force of infection is density dependent,
\begin{equation}
\label{eq:foi}
\lambda_i(t) = \beta \sum_{j=1}^{K} W_{ij}\, I_j(t),
\end{equation}
where $\beta > 0$ is a per-pair transmission rate and $W = (W_{ij})$ is the Who-Acquires-Infection-From-Whom (WAIFW) matrix. Following \citet{anderson1985agerelated,shea2014adaptive}, $W_{ij}$ depends only on $|i-j|$, decreasing from $1$ on the diagonal to $2/3$ for $|i-j|\ge 4$ (the full specification is given in Section~\ref{sec:si-waifw} of the Supplementary Material), encoding contact rates that are highest within an age class and decline with age separation.

At $t = 0$, the susceptible age distribution is a discretized Gamma density: $S_i(0)$ is proportional to the $\mathrm{Gamma}(a_{\mathrm{sh}}, a_{\mathrm{rt}})$ density at the class midpoint, normalized so that $\sum_{i=1}^{K} S_i(0) = N_0$, the pre-outbreak susceptible population size. A small fraction $f_E \in (0,1)$ of each class is moved into the exposed compartment to seed the epidemic, $E_i(0) = \max\{f_E \cdot S_i(0),\, 1\}$, with $S_i(0)$ adjusted accordingly and $I_i(0) = R_i(0) = 0$. Because the elapsed time between epidemic onset and the first reporting week is unknown, we introduce a warm-up duration $t_0 > 0$: the system is integrated from $t=0$ to $t=t_0$, and the simulator is then evaluated at $t=t_0+7(u-1)$ days for $u=1,\ldots,T$. The resulting weekly case-count sequence is the simulator output $Y(\boldsymbol{\theta}) \in \mathbb{R}^{T}$, and the time-summed age-group proportions of the infectious compartment over the observation window,
\begin{equation*}
a_g = \frac{\sum_{i \in \mathcal{G}_g} \int_{t_0}^{t_0 + 364} I_i(t)\, dt}{\sum_{i=1}^{K} \int_{t_0}^{t_0 + 364} I_i(t)\, dt}, \qquad g \in \{1,2,3\},
\end{equation*}
with $\mathcal{G}_g$ the index sets of the three age groups, form the auxiliary summary $A(\boldsymbol{\theta}) = (a_1, a_2, a_3) \in \Delta^{2}$. The calibration target is the six-dimensional vector $\boldsymbol{\theta} = (\beta,\, N_0,\, f_E,\, a_{\mathrm{sh}},\, a_{\mathrm{rt}},\, t_0)^{\top}$, sampled from a Latin hypercube design with the bounds in Table~\ref{tab:seir-priors}.

\begin{table}[!tbp]
\centering
\begin{tabular}{lll}
\toprule
Parameter & Definition & Prior support \\
\midrule
$\beta$                  & Per-pair transmission rate (day$^{-1}$)  & $[0.5 \times 10^{-6},\ 2.5 \times 10^{-6}]$ \\
$N_0$                    & Susceptible population size              & $[1.4 \times 10^{5},\ 3 \times 10^{5}]$ \\
$f_E$                    & Initial exposed fraction                 & $[10^{-4},\ 10^{-3}]$ \\
$a_{\mathrm{sh}}$ & Shape of age distribution             & $[0.8,\ 1.4]$ \\
$a_{\mathrm{rt}}$ & Rate of age distribution              & $[0.09,\ 0.39]$ \\
$t_0$                   & Warm-up duration (days)                  & $[0.1,\ 10]$ \\
\bottomrule
\end{tabular}
\caption{Latin hypercube bounds for the SEIR calibration parameters.}
\label{tab:seir-priors}
\end{table}

\subsection{Discrepancy, Synthetic Data, and Implementation}
\label{subsec:SEIR-implementation-details}

We assume that the discrepancy between $Y(\boldsymbol{\theta})$ and the observed weekly cases is generated by the heterogeneous GP family of Section~\ref{subsec:discrepancy}, with hyperparameters sampled via LHS from bounds calibrated to the scale of the measles data: $\tau_{\mathrm{prop}} \in [0.1,\, 0.3]$, $\kappa \in [0.10^2,\, 0.40^2]$, and $\zeta \in [0.01^2,\, 0.04^2]$; $A(\boldsymbol{\theta})$ is not subject to discrepancy. The training set is constructed exactly as in Section~\ref{subsec:training-data-generation}, with $N_{\mathrm{train}} = 5{,}000$ LHS parameter configurations from Table~\ref{tab:seir-priors}, $N_{\mathrm{noise}} = 30$ GP hyperparameter vectors, no state-dependent envelope, and $150{,}000$ contaminated training sequences. The CRC calibration dataset $\mathcal{D}_n$ of size $n = 100$ follows Section~\ref{subsec:calibration-data-generation}: parameters are drawn through the trained flow map conditioned on $(A_{\mathrm{obs}}, Z_{\mathrm{obs}})$, clamped to the support in Table~\ref{tab:seir-priors}, passed through the SEIR model, and contaminated with a unique GP discrepancy modulated by the envelope of equation~\eqref{eq:envelope}, giving independent draws from equation~\eqref{eq:joint-distribution}. Under the working assumption that the real-world discrepancy is itself a realization from this heteroscedastic GP family, $\mathcal{D}_n$ and the target observation are exchangeable.

The flow-matching architecture, sampling ODE, parameter-set constructions, and evaluation metrics are identical to those of Section~\ref{subsec:toy-implementation-details}. The signal-relevance threshold is $\eta = 300$ weekly cases, the level below which apparent miscoverage carries little epidemiological consequence, and the tuning procedure of Section~\ref{subsec:rho-selection} selects $\rho_{\mathrm{rad}}=0.03$ and $\rho_{\mathrm{pos}}=0.05$. The thresholds $\widehat{\lambda}_{\mathrm{rad}}$ and $\widehat{\lambda}_{\mathrm{pos}}$ are calibrated once on $\mathcal{D}_n$ and held fixed. As a baseline, KOH calibration is run by MCMC for the real observation to sample from the posterior predictive distribution, from which pointwise $(1-\alpha)$ credible intervals are constructed; MCMC specifications and convergence diagnostics are given in Section~\ref{sec:si-mcmc-real} of the Supplementary Material.

\subsection{Results}

\begin{figure}[!tbp]
\centering
\begin{subfigure}{0.8\textwidth}
\centering
\includegraphics[width=\linewidth]{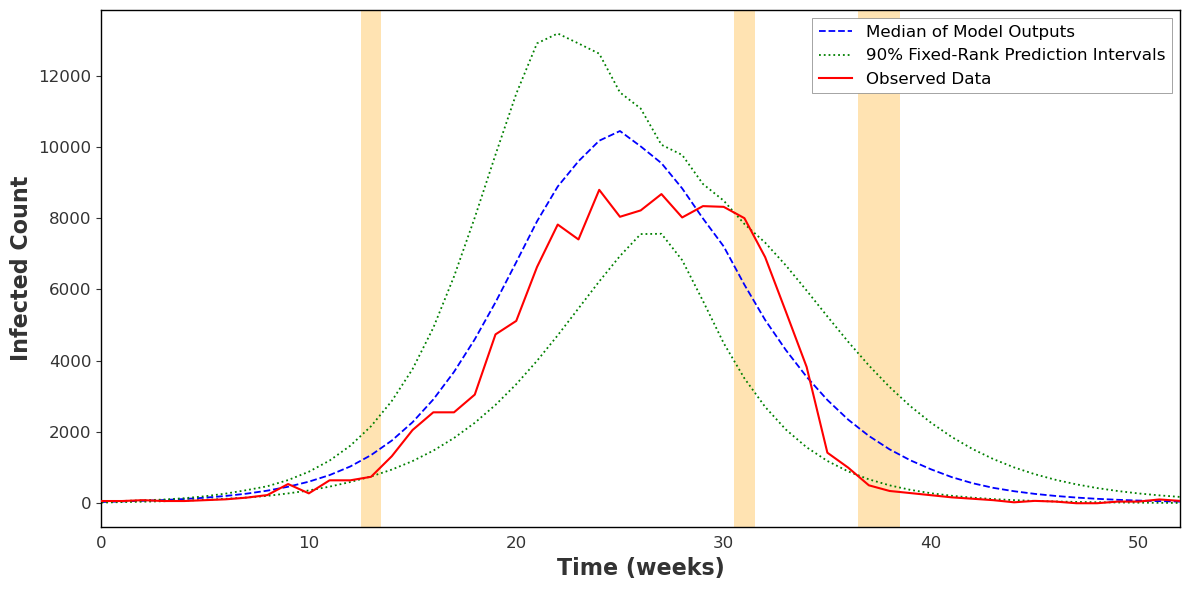}
\caption{Radius-based construction.}
\label{fig:seir-radius}
\end{subfigure}

\smallskip
\begin{subfigure}{0.8\textwidth}
\centering
\includegraphics[width=\linewidth]{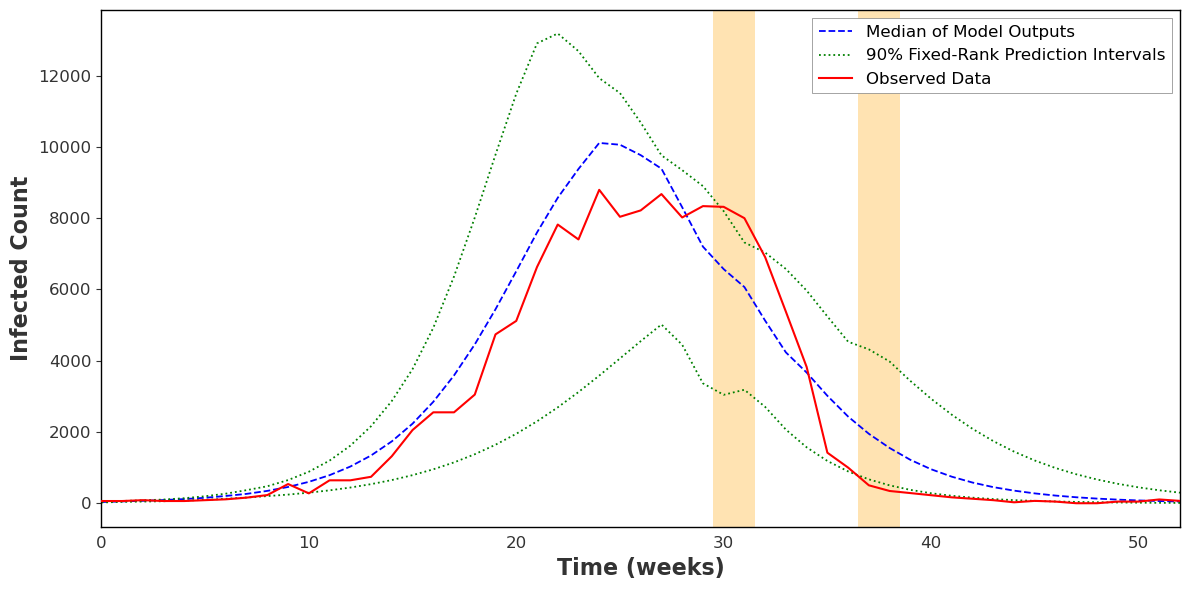}
\caption{Posterior-based construction.}
\label{fig:seir-likelihood}
\end{subfigure}
\caption{CRC prediction intervals for the Malawi measles outbreak. The blue dashed line is the median of the generated SEIR trajectories, the green dotted lines are the pointwise prediction bounds, and the red solid line is the observed series. Orange bands mark weeks at which the observed value falls outside the prediction interval and exceeds the signal-relevance threshold $\eta = 300$.}
\label{fig:seir-crc-intervals}
\end{figure}

Table~\ref{tab:seir-results} reports empirical miscoverage, RMSE between the median predicted trajectory and the observed weekly cases, and average pointwise interval length for the two proposed constructions and the MCMC baseline. Figure~\ref{fig:seir-crc-intervals} displays the proposed bands overlaid on the Malawi data, with miscovered weeks highlighted in orange; the KOH--MCMC band was shown in Figure~\ref{fig:seir-mcmc}.

\begin{table}[!tbp]
\centering
\begin{tabular}{lccc}
\toprule
Method & Miscoverage & RMSE & Avg. interval length \\
\midrule
Radius-based      & 0.1379 & 940.77 & 2540.26 \\
Posterior-based  & 0.1379 & 908.40 & 3157.48 \\
MCMC baseline     & 0.0345 & 947.96  & 3526.89 \\
\midrule
Target ($\alpha=0.1$) & 0.1000 & --- & --- \\
\bottomrule
\end{tabular}
\caption{Empirical miscoverage, RMSE between the median predicted weekly cases and the observed measles outbreak, and average pointwise interval length for the radius-based parameter set, posterior-based parameter set, and MCMC baseline. The target miscoverage level is $\alpha = 0.1$.}
\label{tab:seir-results}
\end{table}

Both proposed parameter sets achieve an empirical miscoverage of $0.1379$, moderately above the target $\alpha = 0.1$ and consistent with the CRC guarantee $\mathbb{E}[L_{n+1}(\hat\lambda)] \le \alpha$, with average interval lengths of $2540.26$ and $3157.48$ cases. The MCMC baseline attains a miscoverage of $0.0345$, less than half the nominal level, with the widest band ($3526.89$ cases); this pattern is consistent with the sensitivity of posterior uncertainty to the assumed discrepancy specification \citep{brynjarsdottir2014learning,syring2019calibrating}. The CRC procedure instead calibrates $\hat\lambda$ directly against the empirical miscoverage of synthetic data generated under the heterogeneous GP family, sidestepping the need for any single parametric noise model to be exactly correct. The RMSE values of the three methods are comparable. The ablation study in Section~\ref{sec:ablation-study} of the Supplementary Material isolates the role of the CRC step on these data: retaining a fixed $90\%$ of the generated draws without CRC, under either selection rule, attains zero miscoverage on the evaluated weeks but produces bands considerably wider than both the CRC and KOH--MCMC bands, indicating overcoverage (cf.\ Figure~\ref{fig:naive-sbi}).

A further feature of the framework is that the miscovered weeks, shown as orange bands in Figure~\ref{fig:seir-crc-intervals}, localize where the SEIR model struggles most. In both CRC panels the orange bands cluster around the peak and in the late tail, regions corresponding to the asymmetric falloff of the real outbreak---a pattern that the approximately symmetric, bell-shaped SEIR trajectories cannot reproduce regardless of parameter choice. The CRC framework therefore supplies not only a coverage guarantee but also a model-diagnostic signal: when the calibrated interval fails, the model is structurally inadequate. The MCMC procedure offers no analogous diagnostic because its credible intervals do not target coverage; its overcoverage reflects the same structural inadequacy but provides no statistical mechanism for surfacing it. In applications where the validity of the predictive interval is itself of scientific interest, this combination of finite-sample guarantees and built-in diagnostics supports the proposed approach.

\section{Discussion}
\label{sec:discussion}

This paper has developed a framework for computer model calibration that combines amortized neural posterior approximation via flow matching with conformal risk control to deliver finite-sample miscoverage guarantees on the simulator pushforward of posterior samples. In both a synthetic benchmark with known ground truth and a real-data calibration of an age-structured SEIR model to the 2010 Malawi measles outbreak, the proposed parameter sets attained near-nominal coverage with substantially sharper bands, whereas a standard KOH--MCMC baseline overcovered. The contrast reflects a difference in objective rather than in implementation quality: a Bayesian analysis under a Gaussian observation model reports credible intervals as a byproduct of posterior sampling and has limited capacity to account for structural mismatch between a mechanistic model and reality, whereas our framework targets predictive miscoverage directly and narrows the parameter set only as much as the calibration data demand.

Two features make the framework well-suited to applied calibration. First, it is \emph{amortized}: the flow-matching posterior is trained once and queried rapidly for any new observation, in contrast to MCMC, which must be rerun each time \citep{cranmer2020frontier,dax2023flowmatching}---a consequential advantage for real-time surveillance and sequential calibration where data accrue continuously. More fundamentally, amortization is what makes the CRC layer possible at all: the calibration and tuning samples together require on the order of $n+n_{\mathrm{tune}}+n_v$ conditional posterior approximations per target observation, a cost that is negligible for a trained flow but prohibitive for observation-specific MCMC. Second, it supplies an interpretable \emph{diagnostic}: clusters of time points at which the calibrated interval fails to cover localize the regions where the mechanistic model is structurally inadequate, directing attention toward the aspects of the model most in need of refinement.

Several limitations should be acknowledged. The finite-sample guarantee relies on exchangeability between the synthetic calibration set and the real observation \citep{angelopoulos2024conformal}, which holds under the working assumption that the true discrepancy lies in the user-specified GP family---a condition that cannot be verified from a single observation. The quality of the amortized posterior depends on the size and coverage of the training simulation pool, and higher-dimensional parameter spaces may require larger training budgets \citep{cranmer2020frontier,lueckmann2021benchmark}. Finally, the framework controls marginal miscoverage averaged over time; pointwise or conditional miscoverage at specific time points is not guaranteed.

Several extensions suggest themselves. Although exchangeability is untestable in the strict sense, partial diagnostics comparing summary statistics of the real observation against their empirical distribution across the calibration set---for instance, a kernel two-sample test on residual sequences \citep{gretton2012kernel}---would help flag gross violations. Mondrian conformal prediction \citep{vovk2005algorithmic} or localized conformal variants \citep{guan2023localized} could deliver control that adapts to local difficulty, a multi-channel extension would accommodate several observation channels with distinct discrepancy structures, and moving beyond the fixed-kernel GP family to non-stationary or non-Gaussian discrepancies, or learning the discrepancy family adaptively, would improve robustness under misspecification. More broadly, the calibration literature has long recognized discrepancy between model and reality as the principal source of inferential difficulty \citep{kennedy2001bayesian,brynjarsdottir2014learning,tuo2015efficient}. The framework presented here identifies the region of parameter space most relevant for prediction by calibrating prediction intervals against a flexible family of discrepancies, and we hope it proves useful as a robust complement to classical KOH calibration for scientific problems in which mechanistic models meet imperfect data.

\ifblind\else
\vspace{-5pt}
\section*{Acknowledgments}
\vspace{-5pt}
{\raggedright This work was supported by the National Research Foundation of Korea (NRF) grant funded by the Ministry of Science and ICT (MSIT) (grant number [RS-2025-00523567,RS-2026-25607210]), the Global-LAMP Program of the NRF grant funded by the Ministry of Education of the Republic of Korea (grant number [RS-2023-00301976]).\par}
\fi

\vspace{-5pt}
\section*{Supplementary Material}
\vspace{-5pt}
Supplementary material available online contains the full flow-matching architecture, training details, and a general formulation of flow matching (Section~\ref{sec:si-model-details}), the WAIFW matrix (Section~\ref{sec:si-waifw}), MCMC specifications and diagnostics (Section~\ref{sec:mcmc-specifications}), and an ablation study without conformal risk control (Section~\ref{sec:ablation-study}). \ifblind The source code will be made publicly available upon publication.\else The source code can be downloaded from \href{https://github.com/ryanb61901/Prediction-Uncertainty-Controlled-Computer-Model-Calibration}{github.com/ryanb61901/Prediction-Uncertainty-Controlled-Computer-Model-Calibration}.\fi

\vspace{-5pt}
\section*{Data Availability Statement}
\vspace{-5pt}
The data that support the findings of this study are available from the authors upon reasonable request.

\bibliography{Reference.bib}

\ifarxiv
  % ---- arXiv version: append the supplement (supplement_body.tex) ----
  \clearpage
  \setcounter{section}{0}\setcounter{equation}{0}%
  \setcounter{figure}{0}\setcounter{table}{0}%
  \renewcommand{\thesection}{S\arabic{section}}%
  \renewcommand{\theequation}{S\arabic{equation}}%
  \renewcommand{\thefigure}{S\arabic{figure}}%
  \renewcommand{\thetable}{S\arabic{table}}%
  % unique hyperlink anchors for the S-numbered objects
  \renewcommand{\theHsection}{S\arabic{section}}%
  \renewcommand{\theHequation}{S\arabic{equation}}%
  \renewcommand{\theHfigure}{S\arabic{figure}}%
  \renewcommand{\theHtable}{S\arabic{table}}%
  \begin{center}{\LARGE\bfseries Supplementary Material}\end{center}
  \bigskip
  % ===========================================================================
% Body of the Supplementary Material (no preamble, no \begin{document}).
% Edit the supplement's content HERE. This file is \input by:
%   - Supplementary.tex  (journal version: separate PDF with its own title page)
%   - main.tex when \arxivtrue  (arXiv version: one combined PDF)
% ===========================================================================

\section{Model Details}\label{sec:si-model-details}

\subsection{Flow-Matching Architecture}\label{sec:si-architecture}

This section gives the complete specification of the network summarized in Section~\ref{subsec:flow-matching-network} of the main text. To parameterize the conditional posterior of the calibration parameters, we utilize a deep neural network to approximate the target conditional vector field $v_t^\phi(x\mid A,Z)$ \citep{dax2023flowmatching,lipman2023flowmatching,lipman2024flowmatchingguidecode}. Let $\theta\in\mathbb{R}^d$ denote the target calibration parameter and $x\in\mathbb{R}^d$ the continuous latent state at integration time $t\in[0,1]$. The conditional vector field is defined as a mapping
\begin{equation*}
v_t^\phi(\cdot\mid \cdot,\cdot): \mathbb{R}^d \times \mathcal{A} \times \mathbb{R}^T \times [0,1] \to \mathbb{R}^d,\ (x, A, Z, t)\mapsto v_t^\phi(x\mid A,Z)
\end{equation*}
governed by trainable network weights $\phi$. In practice, we parameterize $v_t^\phi(x\mid A,Z)$ as a neural network with inputs $(x,A,Z,t)$.

Let $a=\dim(A)$, with $a=0$ when no auxiliary covariates are available. We inject information on $(t,A,Z)$ through the context vector
\begin{equation*}
c_t=[h_t,\,h_Z,\,A]\in\mathbb{R}^{r_c},
\end{equation*}
where $h_t\in\mathbb{R}^{r_t}$ is the time embedding and $h_Z$ is the sequence representation defined below.

Let $\omega\in\mathbb{R}^{r_t/2}$ be a vector of frequencies sampled once from $\mathcal{N}(0,\sigma_\omega^2I_{r_t/2})$ and then held fixed \citep{rahimi2007randomfeatures}. Define
\begin{equation*}
r(t)=\sqrt{2}\,[\sin(2\pi t\omega)^\top,\cos(2\pi t\omega)^\top]^\top\in\mathbb{R}^{r_t},
\end{equation*}
and pass it through the two-layer SiLU network
\begin{equation*}
h_t=W_{t,2}\,\mathrm{SiLU}(W_{t,1}r(t)+b_{t,1})+b_{t,2}\in\mathbb{R}^{r_t},
\end{equation*}
where $W_{t,1}\in\mathbb{R}^{2r_t\times r_t}$, $W_{t,2}\in\mathbb{R}^{r_t\times2r_t}$, $b_{t,1}\in\mathbb{R}^{2r_t}$, and $b_{t,2}\in\mathbb{R}^{r_t}$. In our implementation, $r_t=256$ and $\sigma_\omega=16$ \citep{lipman2024flowmatchingguidecode}.

To construct $h_Z$, we first divide $Z\in\mathbb{R}^T$ into $L=\lfloor(T-p)/s\rfloor+1$ overlapping patches $P_\ell(Z)\in\mathbb{R}^{p}$ of width $p=7$ and stride $s=2$. Each patch is embedded as $e_\ell=W_{\mathrm{patch}}P_\ell(Z)+b_{\mathrm{patch}}\in\mathbb{R}^{r_Z}$, where $W_{\mathrm{patch}}\in\mathbb{R}^{r_Z\times p}$, $b_{\mathrm{patch}}\in\mathbb{R}^{r_Z}$, and $r_Z=128$. After a learned start-of-sequence token in $\mathbb{R}^{r_Z}$ is prepended and positional encodings are added, a four-layer Transformer encoder maps the sequence to
\begin{equation*}
H(Z)\in\mathbb{R}^{(L+1)\times r_Z}.
\end{equation*}

We aggregate $H(Z)$ using learned-query multihead-attention pooling \citep{vaswani2017attention,lee2019settransformer}. Let $Q_{\mathrm{pool}}\in\mathbb{R}^{q\times r_Z}$ contain $q=4$ learned query vectors, let $H_a=4$ be the number of attention heads, and let $r_h^{\mathrm{att}}=r_Z/H_a=32$ be the dimension of each head. For head $j\in\{1,\ldots,H_a\}$, define
\begin{equation*}
O_j=\operatorname{softmax}\!\left(\frac{(Q_{\mathrm{pool}}W_{Q,j})(H(Z)W_{K,j})^\top}{\sqrt{r_h^{\mathrm{att}}}}\right)H(Z)W_{V,j}\in\mathbb{R}^{q\times r_h^{\mathrm{att}}},
\end{equation*}
where $W_{Q,j},W_{K,j},W_{V,j}\in\mathbb{R}^{r_Z\times r_h^{\mathrm{att}}}$ are learned query, key, and value projections. The head outputs are concatenated columnwise and projected as
\begin{equation*}
O=\operatorname{Concat}(O_1,\ldots,O_{H_a})W_O+\boldsymbol{1}_q b_O^\top\in\mathbb{R}^{q\times r_Z},
\end{equation*}
where $W_O\in\mathbb{R}^{r_Z\times r_Z}$, $b_O\in\mathbb{R}^{r_Z}$, and $\boldsymbol{1}_q$ is the length-$q$ vector of ones. For a matrix $O$, $\operatorname{Flatten}(O):=(O_{1,:}^\top,\ldots,O_{q,:}^\top)^\top\in\mathbb{R}^{qr_Z}$ denotes rowwise vectorization. The pooled sequence embedding is therefore
\begin{equation*}
h_{\mathrm{seq}}=W_{\mathrm{proj}}\operatorname{Flatten}(O)+b_{\mathrm{proj}}\in\mathbb{R}^{r_Z},
\end{equation*}
with $W_{\mathrm{proj}}\in\mathbb{R}^{r_Z\times qr_Z}$ and $b_{\mathrm{proj}}\in\mathbb{R}^{r_Z}$.

We augment this learned representation with two geometric features. Let $u^*=\arg\max_{u\in\{1,\ldots,T\}}Z_u$, and define $\tilde Z_{\max}=\log\{1+\max_u Z_u\}$ and $\tilde Z_{\arg}=(u^*-1)/T$, where the subtraction by one matches the zero-based index used in the implementation. Then
\begin{equation*}
h_Z=[h_{\mathrm{seq}},\tilde Z_{\max},\tilde Z_{\arg}]\in\mathbb{R}^{r_Z+2},\qquad
r_c=r_t+r_Z+2+a.
\end{equation*}

Given $x$ and $c_t$, the vector-field backbone uses $N_{\mathrm{blk}}=4$ fixed-width residual blocks with hidden dimension $r_b=256$, Adaptive Layer Normalization (AdaLN), and Swish-Gated Linear Units (SwiGLU) \citep{shazeer2020glu,peebles2023scalable}. The initial state is $h^{(0)}=W_xx+b_x\in\mathbb{R}^{r_b}$, where $W_x\in\mathbb{R}^{r_b\times d}$ and $b_x\in\mathbb{R}^{r_b}$. In block $m\in\{1,\ldots,N_{\mathrm{blk}}\}$, the context-dependent scale and shift are
\begin{equation*}
\begin{bmatrix}\gamma_m(c_t)\\ \beta_m(c_t)\end{bmatrix}
=W_{c,m}\mathrm{SiLU}(c_t)+b_{c,m}\in\mathbb{R}^{2r_b},
\end{equation*}
where $W_{c,m}\in\mathbb{R}^{2r_b\times r_c}$ and $b_{c,m}\in\mathbb{R}^{2r_b}$. The block then computes
\begin{align*}
\tilde h^{(m)}&=\mathrm{LayerNorm}(h^{(m-1)})\odot\{1+\gamma_m(c_t)\}+\beta_m(c_t),\\
u_m&=W_{\mathrm{val},m}\tilde h^{(m)}+b_{\mathrm{val},m},\qquad
g_m=W_{\mathrm{gate},m}\tilde h^{(m)}+b_{\mathrm{gate},m},\\
h^{(m)}&=h^{(m-1)}+W_{\mathrm{out},m}\mathrm{Dropout}\!\{u_m\odot\mathrm{SiLU}(g_m)\}+b_{\mathrm{out},m},
\end{align*}
where $W_{\mathrm{val},m},W_{\mathrm{gate},m},W_{\mathrm{out},m}\in\mathbb{R}^{r_b\times r_b}$, their corresponding biases lie in $\mathbb{R}^{r_b}$, and $\odot$ denotes elementwise multiplication. Here LayerNorm, Dropout, and SiLU denote the standard layer-normalization, dropout, and sigmoid-weighted linear-unit operations, respectively. Finally, $\mathrm{LayerNorm}(h^{(N_{\mathrm{blk}})})$ is concatenated with $c_t$ and passed through $d$ coordinate-specific output heads $g_1^\phi,\ldots,g_d^\phi$, giving
\begin{equation*}
v_t^\phi(x\mid A,Z)=\big(g_1^\phi([\mathrm{LayerNorm}(h^{(N_{\mathrm{blk}})}),c_t]),\ldots,g_d^\phi([\mathrm{LayerNorm}(h^{(N_{\mathrm{blk}})}),c_t])\big)^\top\in\mathbb{R}^d.
\end{equation*}
The context projections and $W_{\mathrm{out},m}$ are initialized to zero, so each residual block initially acts as the identity. The output heads are separately initialized to zero, which makes the initial vector-field output $v_t^\phi$ equal to zero and improves numerical stability early in training.

The sequence encoder is a Transformer with $4$ layers, $4$ attention heads, feedforward dimension $512$, and dropout rate $0.1$ \citep{vaswani2017attention}. The model is trained on the $150{,}000$ contaminated trajectories using the conditional flow-matching objective with $\sigma_{\mathrm{min}}=0.1$.

\subsection{Flow Matching Details}\label{sec:si-flow-matching}

This section summarizes the general flow-matching framework used in the main text. Flow matching learns a continuous-time transport from a simple reference distribution $p_{\mathrm{init}}$ to a target distribution $p_{\mathrm{data}}$ by regressing a time-dependent vector field \citep{lipman2023flowmatching}. Let $p_t(x\mid z)$ be a conditional probability path with endpoints $p_0(\cdot\mid z)=p_{\mathrm{init}}$ and $p_1(\cdot\mid z)=\mathcal{N}(z,\sigma_\mathrm{min}^2I)\approx\delta_z$. A conditional vector field $v_t^{\mathrm{flow}}(\cdot\mid z)$ generates this path if the solution of
\begin{equation*}
dX_t=v_t^{\mathrm{flow}}(X_t\mid z)dt,
\end{equation*}
with $X_0\sim p_0(\cdot\mid z)$ satisfies $X_t\sim p_t(\cdot\mid z)$ for each $t\in[0,1]$.

Integrating over $z\sim p_{\mathrm{data}}$ yields the marginal path
\begin{equation*}
p_t(x):=\int p_t(x\mid z)p_{\mathrm{data}}(z)\,dz,
\end{equation*}
with endpoints $p_0=p_{\mathrm{init}}$ and $p_1=\mathcal{N}(0,\sigma_\text{min}^2I)*p_{\mathrm{data}}\approx p_{\mathrm{data}}$. The corresponding marginal vector field is
\begin{equation*}
v_t^{\mathrm{target}}(x)=\int v_t^{\mathrm{flow}}(x\mid z)\frac{p_t(x\mid z)p_{\mathrm{data}}(z)}{p_t(x)}\,dz,
\end{equation*}
which drives
\begin{equation*}
X_0\sim p_{\mathrm{init}},\qquad dX_t=v_t^{\mathrm{target}}(X_t)dt,\qquad X_1\dot{\sim} p_{\mathrm{data}}.
\end{equation*}

The ideal objective is
\begin{equation*}
\mathcal{L}_{\mathrm{FM}}(\phi)=\mathbb{E}_{t\sim \pi,\,x\sim p_t}\left[\|v_t^{\phi}(x)-v_t^{\mathrm{target}}(x)\|^2\right],
\end{equation*}
but $v_t^{\mathrm{target}}$ is often intractable. In practice, one minimizes the conditional objective
\begin{equation*}
\mathcal{L}_{\mathrm{CFM}}(\phi)=\mathbb{E}_{t\sim \pi,\,z\sim p_{\mathrm{data}},\,x\sim p_t(\cdot\mid z)}\left[\|v_t^{\phi}(x)-v_t^{\mathrm{flow}}(x\mid z)\|^2\right],
\end{equation*}
which has the same gradient target under standard regularity conditions, i.e., $\nabla_\phi\mathcal{L}_{\mathrm{CFM}}(\phi)=\nabla_\phi\mathcal{L}_{\mathrm{FM}}(\phi)$ \citep{lipman2023flowmatching,lipman2024flowmatchingguidecode}.

After training, approximate samples are obtained by solving
\begin{equation*}
X_0\sim p_{\mathrm{init}},\qquad dX_t=v_t^{\phi}(X_t)dt,\qquad t\in[0,1],
\end{equation*}
and taking $X_1$.

For Gaussian probability paths,
\begin{equation*}
p_t(x\mid z)=\mathcal{N}(x;\alpha_t z,\beta_t^2 I_d),
\end{equation*}
with smooth schedules $(\alpha_t,\beta_t)$ satisfying $\alpha_0=0$, $\alpha_1=1$, $\beta_0=1$, and $\beta_1=\sigma_{\min}$, the conditional velocity is
\begin{equation*}
v_t^{\mathrm{flow}}(x\mid z)=\left(\dot\alpha_t-\frac{\dot\beta_t}{\beta_t}\alpha_t\right)z+\frac{\dot\beta_t}{\beta_t}x.
\end{equation*}
% If $s_t(x):=\nabla_x\log p_t(x)$ denotes the score, then
% \begin{equation*}
% v_t^{\mathrm{target}}(x)=a_t s_t(x)+b_t x,\qquad a_t=\beta_t^2\frac{\dot\alpha_t}{\alpha_t}-\beta_t\dot\beta_t,\qquad b_t=\frac{\dot\alpha_t}{\alpha_t},
% \end{equation*}
% so the score is recoverable from a learned velocity:
% \begin{equation*}
% s_t^{\phi}(x)=\frac{v_t^{\phi}(x)-b_t x}{a_t}.
% \end{equation*}
% This gives the SDE sampler
% \begin{equation*}
% dX_t=\left[v_t^{\phi}(X_t)+\frac{\sigma_t^2}{2}s_t^{\phi}(X_t)\right]dt+\sigma_t dW_t,
% \end{equation*}
% where $\sigma_t\ge 0$ controls stochasticity and $W_t$ is a Wiener process \citep{lipman2024flowmatchingguidecode,song2021scorebased}.

In practice, training uses Monte Carlo estimates of the CFM objective. At each update, we draw a minibatch of target--condition pairs $(z_b,y_b)$ from the training data, times $t_b\sim\pi$, and independent noise vectors $\varepsilon_b\sim\mathcal{N}(0,I_d)$; we then form $x_{t_b}=t_bz_b+\{1-(1-\sigma_{\mathrm{min}})t_b\}\varepsilon_b$ and evaluate the closed-form target $v_{t_b}^{\mathrm{flow}}(x_{t_b}\mid z_b)$. The network parameters $\phi$ are updated by backpropagation to minimize the minibatch mean squared error between this target and $v_{t_b}^{\phi}(x_{t_b}\mid y_b)$, thereby learning the conditional vector field over the full probability path without solving an ODE during training.

For conditional generation, guided flows interpolate unconditional and conditional fields:
\begin{equation*}
\tilde v_t(x\mid y)=\bigl(1-w_t\bigr)v_t^{\mathrm{target}}(x\mid \emptyset)+w_tv_t^{\mathrm{target}}(x\mid y),\qquad w_t>1,
\end{equation*}
which increases adherence to the condition $y$ at sampling time \citep{ho2022classifierfree,zheng2023guidedflows}.

\section{WAIFW Matrix for the Age-Structured SEIR Model}\label{sec:si-waifw}

We adopt a structured WAIFW matrix in which the entry between age classes $i$ and $j$ depends only on $|i - j|$, following \citet{anderson1985agerelated} and \citet{shea2014adaptive}: 
\begin{equation*}
W_{ij} =
\begin{cases}
1, & |i - j| = 0,\\
11/12, & |i - j| = 1,\\
5/6, & |i - j| = 2,\\
3/4, & |i - j| = 3,\\
2/3, & |i - j| \geq 4.
\end{cases}
\end{equation*}
This off-diagonal decay encodes the empirical observation that contact rates are highest within an age class and decline gradually with age separation.

\section{MCMC Specifications and Diagnostics}\label{sec:mcmc-specifications}

The conventional Bayesian calibration baselines reported in the main text were
implemented in CmdStan \citep{carpenter2017stan} using the No-U-Turn Sampler (NUTS), an adaptive
Hamiltonian Monte Carlo algorithm \citep{hoffman2014nuts}. For both the simulation study
and the real-data application, we ran four independent chains with
\texttt{max\_treedepth}$ =10$ and \texttt{adapt\_delta}$ =0.85$. The bounded
parameter declarations listed below had no additional prior-density statements
in the Stan programs and therefore induce uniform priors over their respective
supports.

In both analyses, the time-series likelihood was a Gaussian process centered at
the relevant forward-model trajectory, following the KOH discrepancy formulation
\citep{kennedy2001bayesian}. In the notation of the main text,
\begin{equation}
 Z\mid\boldsymbol{\theta},\tau,\kappa,\zeta
 \sim \mathcal{N}_{T}\!\left\{
 Y(\boldsymbol{\theta}),\,
 \boldsymbol{\Sigma}(\boldsymbol{\theta},\tau,\kappa,\zeta)
 \right\},
\label{eq:si-mcmc-gp-likelihood}
\end{equation}
where
\begin{equation}
 \Sigma_{uv}(\boldsymbol{\theta},\tau,\kappa,\zeta)
 =
 \sigma_Y^2(\boldsymbol{\theta})
 \left[
 \zeta\,\mathbb{I}(u=v)
 +\kappa\exp\!\left\{-\frac{|u-v|}{\tau}\right\}
 \right],
 \qquad u,v=1,\ldots,T.
\label{eq:si-mcmc-gp-covariance}
\end{equation}
Here $\tau$ is the correlation range, $\kappa$ is the partial sill, $\zeta$ is
the nugget, and $\sigma_Y^2(\boldsymbol{\theta})$ is the temporal variance of
the forward trajectory, as defined in Section~\ref{subsec:training-data-generation} of the main text. The
covariance hyperparameters had independent priors
$\kappa\sim\operatorname{Inv\text{-}Gamma}(2,1)$ and
$\zeta\sim\operatorname{Inv\text{-}Gamma}(2,1)$, where the second argument is
the scale parameter. The prior support for $\tau$ differed between the two
analyses as specified below.

\subsection{Simulation Study}\label{sec:si-mcmc-simulation}

For the simulation study, the calibration parameters had independent priors
\begin{equation*}
 \theta_1,\theta_2,\theta_3\stackrel{\mathrm{ind}}{\sim}
 \operatorname{Uniform}(0,1),
\end{equation*}
and the GP range parameter had prior
\begin{equation*}
 \tau\sim\operatorname{Uniform}(0.05\times53,\,0.20\times53)
 =\operatorname{Uniform}(2.65,\,10.60).
\end{equation*}
The mean in equation~\eqref{eq:si-mcmc-gp-likelihood} was the Gaussian
density-shaped forward trajectory used in the main text,
\begin{equation*}
 Y(\boldsymbol{\theta},s)
 =0.3+
 \frac{\theta_1+0.3}{\sqrt{2\pi(\theta_3+0.1)}}
 \exp\!\left\{
 -\frac{(s-\theta_2+0.5)^2}{\theta_3+0.1}
 \right\},
\end{equation*}
evaluated at $T=53$ equally spaced values of $s$ on $[-2,2]$.

Each of the four chains was run for $100{,}000$ iterations: the first
$30{,}000$ iterations were discarded as warmup and the remaining $70{,}000$
iterations were retained for posterior inference. Thus, the four chains
contributed $280{,}000$ post-warmup draws in total. Figure~\ref{fig:si-simulation-mcmc-trace}
shows the MCMC diagnostics for a single synthetic observation data: Specifically, the chain-specific marginal posterior densities and post-warmup traces
for the three forward-model parameters and the three GP covariance
hyperparameters. The density curves from the four chains largely overlap, and
the traces show no visually apparent chain-specific drift.

\begin{figure}[p]
\centering
\includegraphics[width=\textwidth]{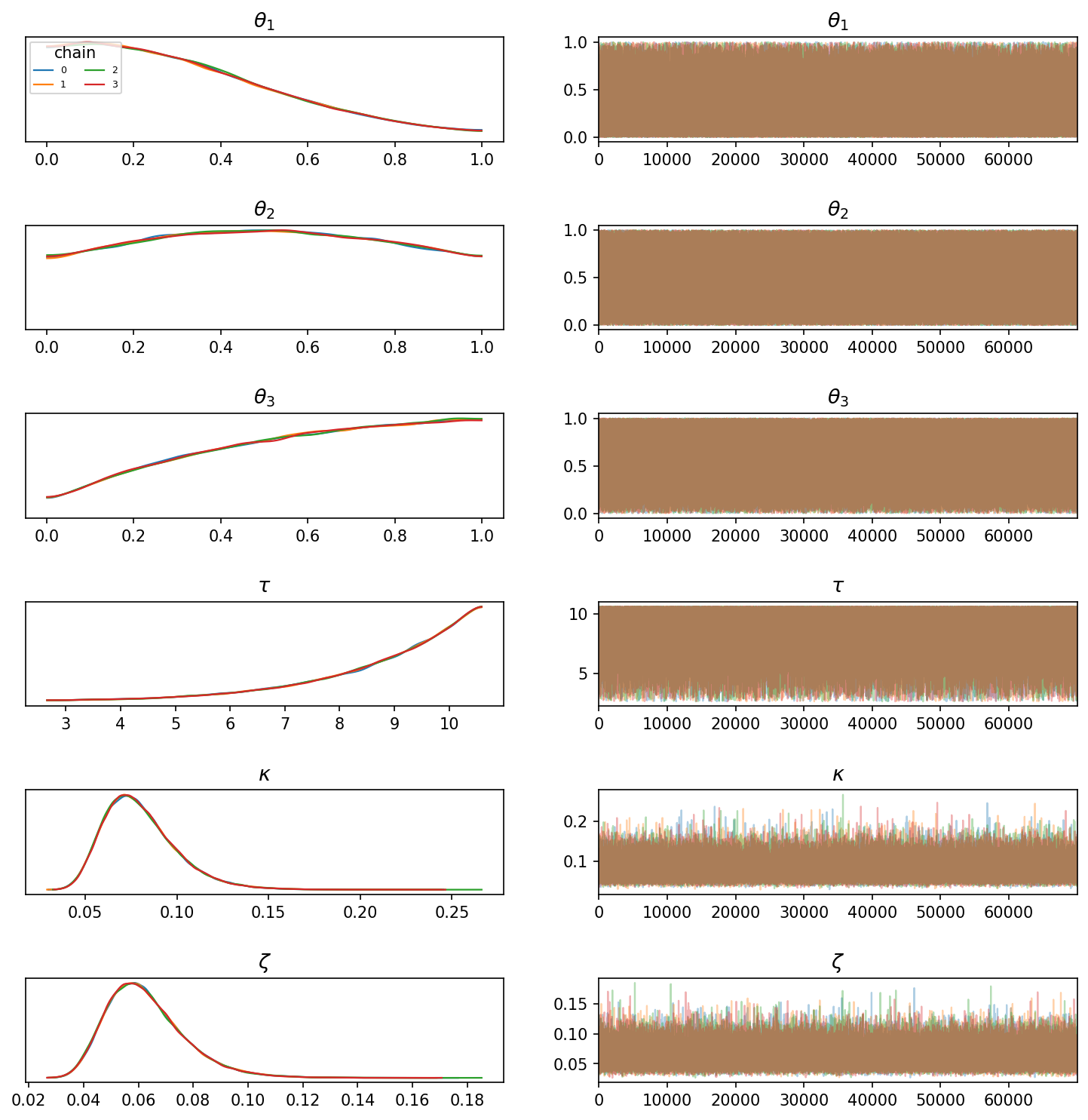}
\caption{MCMC diagnostics for the simulation study. The left column displays
chain-specific marginal posterior density estimates, and the right column
displays the corresponding post-warmup traces for
$\theta_1,\theta_2,\theta_3,\tau,\kappa$, and $\zeta$. Each color represents
one of the four NUTS chains.}
\label{fig:si-simulation-mcmc-trace}
\end{figure}

\subsection{Real-Data Application}\label{sec:si-mcmc-real}

For the measles application, the six forward-model parameters had independent
uniform priors induced by the following bounds:
\begin{equation}
\begin{aligned}
 \beta &\sim \operatorname{Uniform}(0.5\times10^{-6},\,2.5\times10^{-6}),&
 N_0 &\sim \operatorname{Uniform}(1.4\times10^5,\,3.0\times10^5),\\
 f_E &\sim \operatorname{Uniform}(10^{-4},\,10^{-3}),&
 a_{\mathrm{sh}} &\sim \operatorname{Uniform}(0.8,\,1.4),\\
 a_{\mathrm{rt}} &\sim \operatorname{Uniform}(0.09,\,0.39),&
 t_0 &\sim \operatorname{Uniform}(0.1,\,10).
\end{aligned}
\label{eq:si-real-mcmc-priors}
\end{equation}
For this analysis, the GP range parameter had prior
\begin{equation*}
 \tau\sim\operatorname{Uniform}(0.1\times53,\,0.3\times53)
 =\operatorname{Uniform}(5.3,\,15.9).
\end{equation*}

The weekly infected-population time series followed
equations~\eqref{eq:si-mcmc-gp-likelihood}--\eqref{eq:si-mcmc-gp-covariance},
with the deterministic age-structured SEIR trajectory replacing the synthetic
Gaussian density-shaped mean. The age-proportion observations supplied a
second likelihood component. Using the age proportions obtained
from the SEIR trajectory ($A(\boldsymbol{\theta})$), we defined the likelihood of the observed proportions
$A_{\mathrm{obs}}=(0.42,0.30,0.28)^\top$ as
\begin{equation}
 A_{\mathrm{obs}}\mid\boldsymbol{\theta}
 \sim\operatorname{Dirichlet}\!\left\{
 e_2 A(\boldsymbol{\theta})
 \right\},
 \qquad e_2=300.
\label{eq:si-real-age-likelihood}
\end{equation}
The time-series and age-proportion likelihood contributions were conditionally
independent given the forward-model parameters and GP hyperparameters.

Each of the four chains was run for $10{,}000$ iterations, with $3{,}000$
warmup iterations followed by $7{,}000$ retained draws. The combined
post-warmup sample therefore contained $28{,}000$ draws.
Figure~\ref{fig:si-real-mcmc-trace} reports MCMC diagnostics for the observed measles outbreak data: Specifically, the chain-specific marginal
posterior densities and post-warmup traces for the six SEIR parameters and the three GP covariance
hyperparameters. The
four density curves show substantial overlap, and the retained traces show no
obvious long-term chain-specific drift.

\begin{figure}[p]
\centering
\includegraphics[width=0.8\textwidth]{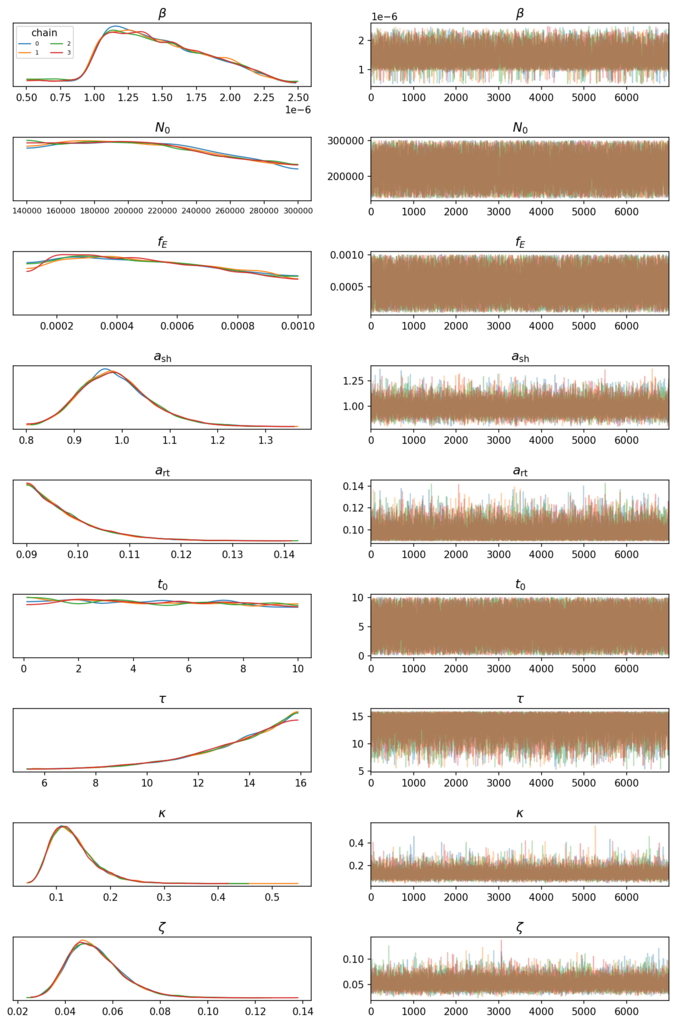}
\caption{MCMC diagnostics for the real-data application. The left column
displays chain-specific marginal posterior density estimates, and the right
column displays the corresponding post-warmup traces for
$\beta,N_0,f_E,a_{\mathrm{sh}},a_{\mathrm{rt}},t_0, \tau,\kappa,$ and $\zeta$. Each color
represents one of the four NUTS chains.}
\label{fig:si-real-mcmc-trace}
\end{figure}

\clearpage

\section{Ablation Study for Parameter-Set Selection}\label{sec:ablation-study}

We conducted an ablation study to examine the empirical behavior of the
prediction intervals when the conformal risk control step is removed. In the
full procedure, the retention threshold is selected by CRC \citep{angelopoulos2024conformal} to control the
expected miscoverage loss. In this ablation, we instead bypass this calibration
step and retain a fixed proportion of candidate draws. Specifically, we retain
the top $90\%$ of candidate draws according to two simple, non-conformal
selection rules. The radius rule retains the draws with the smallest Euclidean
norm of the latent sample, whereas the posterior rule retains the draws with
the largest posterior density values.

After this uncalibrated filtering step, all remaining components of the
pipeline are kept unchanged. The retained draws are propagated through the
simulator, transformed back to the original data scale, clipped to the parameter
bounds, and evaluated using the same fixed-rank empirical miscoverage loss as in
the main analysis. Thus, this ablation isolates the effect of replacing the
CRC-selected parameter set with a fixed $90\%$ parameter set. The resulting
miscoverage values should therefore be interpreted as empirical diagnostics, not
as conformal risk-controlled guarantees.

\begin{table}[htbp]
\centering
\caption{Ablation study without conformal risk control. Instead of selecting
the retention threshold by CRC, each rule retains a fixed $90\%$ of candidate
draws. RMSE is computed between the observed trajectory and the median of the
model outputs, and the average interval length is averaged over time.}
\label{tab:ablation-study}
\begin{tabular}{llccc}
\toprule
Data setting & Filtering rule & Miscoverage & RMSE & Avg. interval length \\
\midrule
Simulated data & Radius & 0.0000 & 0.0119 & 0.0597 \\
Simulated data & Posterior & 0.0000 & 0.0119 & 0.0597 \\
Real data & Radius & 0.0000 & 1041.64 & 6578.77 \\
Real data & Posterior & 0.0000 & 1077.62 & 6078.54 \\
\bottomrule
\end{tabular}
\end{table}

For the simulated data, both uncalibrated filtering rules attain zero empirical
miscoverage, with nearly identical RMSE and average interval length. A similar pattern appears in the
real-data analysis: both rules attain zero empirical miscoverage, with similar RMSE and average interval length.

These results suggest that, in these particular examples, retaining a fixed
$90\%$ of candidate draws produces intervals with empirical miscoverage similar
across the two non-conformal filtering criteria. However, unlike the CRC-based
procedure, this fixed-retention ablation does not adapt the threshold to satisfy
a prescribed risk level and does not provide the finite-sample risk guarantee.
The ablation therefore supports the role of CRC as the component that converts
a reasonable parameter-set construction into a calibrated prediction procedure,
rather than merely serving as another heuristic filtering rule.

\begin{figure}[p]
\centering
\begin{subfigure}{0.94\textwidth}
\centering
\includegraphics[width=\textwidth]{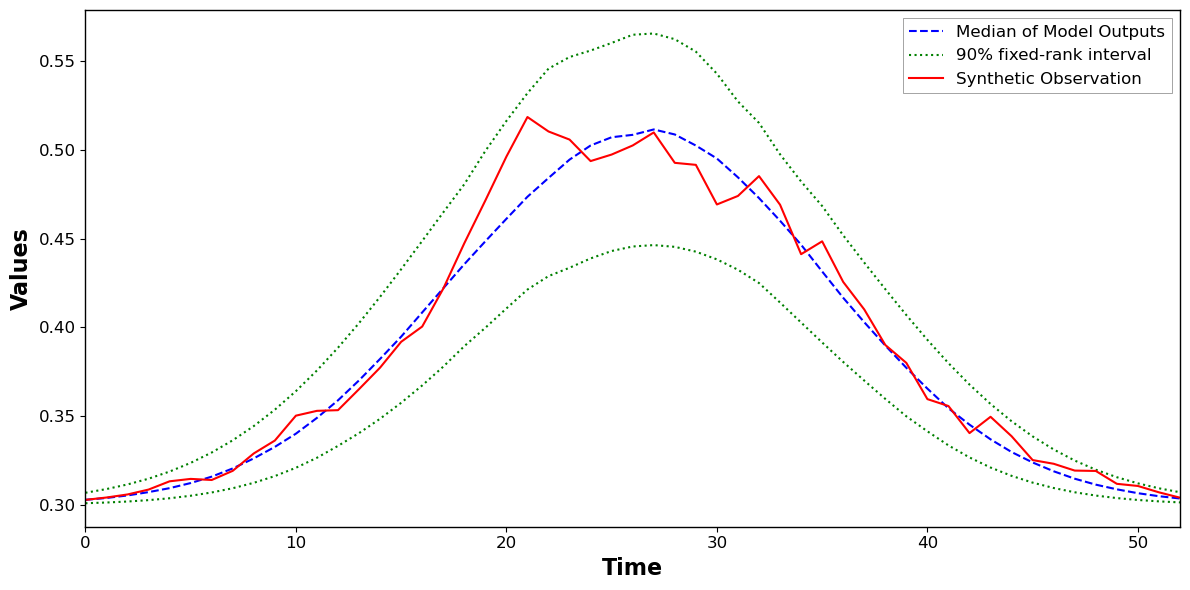}
\caption{Radius-based filtering.}
\label{fig:simulation-radius-ablation}
\end{subfigure}

\vspace{0.4cm}

\begin{subfigure}{0.94\textwidth}
\centering
\includegraphics[width=\textwidth]{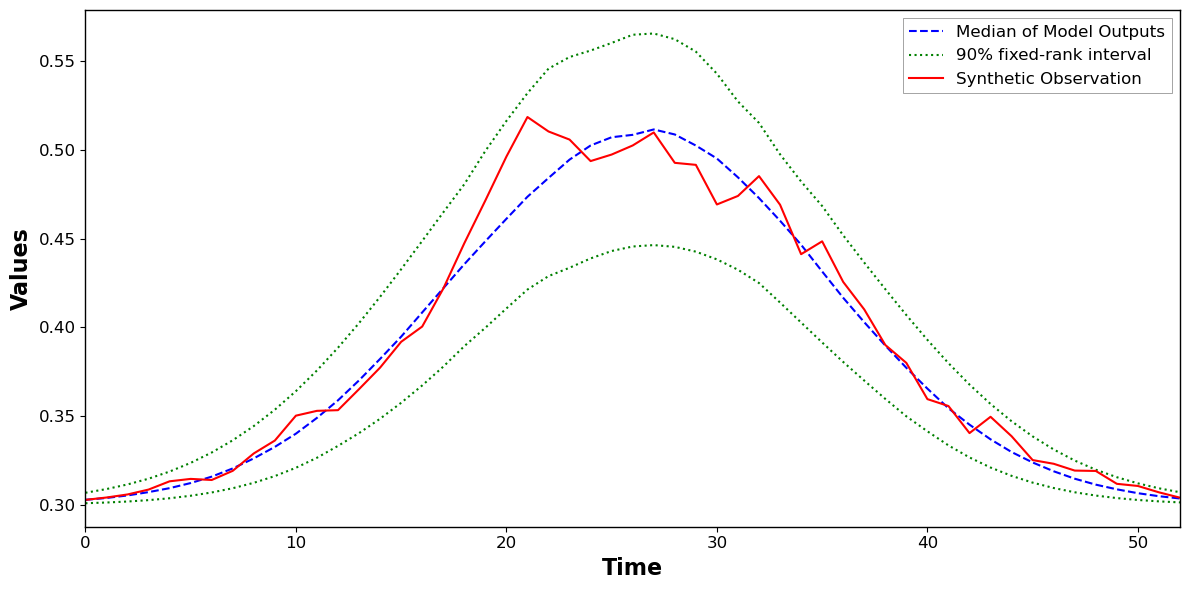}
\caption{Posterior-based filtering.}
\label{fig:simulation-likelihood-ablation}
\end{subfigure}
\caption{Ablation results for the simulated-data experiment. Both panels retain
$90\%$ of the candidate draws and display the median model output, the
$90\%$ fixed-rank prediction interval, and the synthetic observation.}
\label{fig:simulation-ablation}
\end{figure}

\begin{figure}[p]
\centering
\begin{subfigure}{0.94\textwidth}
\centering
\includegraphics[width=\textwidth]{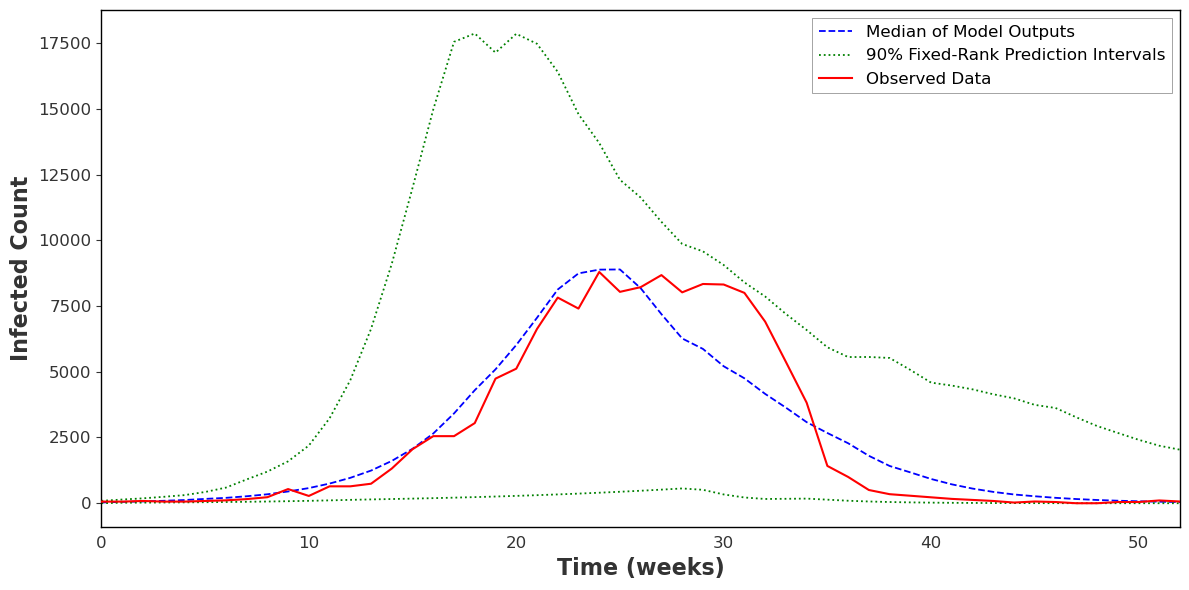}
\caption{Radius-based filtering.}
\label{fig:real-radius-ablation}
\end{subfigure}

\vspace{0.4cm}

\begin{subfigure}{0.94\textwidth}
\centering
\includegraphics[width=\textwidth]{uploads/real_likelihood_ablation.png}
\caption{Posterior-based filtering.}
\label{fig:real-likelihood-ablation}
\end{subfigure}
\caption{Ablation results for the real-data experiment. Both panels retain
$90\%$ of the candidate draws and display the median model output, the
$90\%$ fixed-rank prediction interval, and the observed data.}
\label{fig:real-ablation}
\end{figure}

\fi

\end{document}